\documentclass[journal]{IEEEtran}
\usepackage[citecolor=blue, colorlinks]{hyperref}
\usepackage{graphicx}
\usepackage{float}
\usepackage{subfigure}
\usepackage{epsfig}
\usepackage{epstopdf}
\usepackage{indentfirst}
\usepackage{amsfonts}
\usepackage{amsmath,amssymb}
\usepackage{algorithm}
\usepackage{algorithmic}
\usepackage{verbatim}
\usepackage{multirow}
\usepackage{array}
\usepackage{color}
\usepackage{cite}
\usepackage{makecell}
\usepackage{balance}
\usepackage{hyperref}
\usepackage[table]{xcolor}
\usepackage{fancyhdr}
\usepackage{hyperref}
\usepackage{soul}
\usepackage[normalem]{ulem}
\usepackage{fancyhdr}

\usepackage{bbding}
\usepackage{amssymb}
\usepackage{pifont}
\newcommand{\cmark}{\ding{51}}
\newcommand{\xmark}{\ding{55}}

\ifCLASSINFOpdf
\else
\fi

\begin{document}

\title{Deep Attention-guided Graph Clustering \\with Dual Self-supervision}

% Deep Attention-guided Graph Clustering with Dual Self-supervision

\author{
    Zhihao~Peng,
    Hui~Liu,
	Yuheng~Jia, ~\IEEEmembership{Member,~IEEE}, 
	Junhui~Hou, ~\IEEEmembership{Senior Member,~IEEE}
\thanks{This work was supported in part by the Hong Kong UGC under grant UGC/FDS11/E02/22 and RGC under grants 11219019 and 11202320, in part by the National Natural Science Foundation of China under Grant 62106044, in part by the Natural Science Foundation of Jiangsu Province under Grant BK20210221, in part by the ZhiShan Youth Scholar Program from Southeast University  2242022R40015. Corresponding author: \textit{Hui Liu and Yuheng Jia}.}
\thanks{Z. Peng and J. Hou are with the Department of Computer Science, City University of Hong Kong, Kowloon, Hong Kong 999077 (e-mail: zhihapeng3-c@my.cityu.edu.hk; jh.hou@cityu.edu.hk)}
\thanks{H. Liu is with the School of Computing \& Information Sciences, Caritas Institute of Higher Education, Hong Kong. E-mail:hliu99-c@my.cityu.edu.hk}
\thanks{Y. Jia is with the School of Computer Science and Engineering, Southeast University, Nanjing 210096, China, and also with Key Laboratory of Computer Network and Information Integration (Southeast University), Ministry of Education, China (e-mail: yhjia@seu.edu.cn).}
}

\markboth{IEEE TRANSACTIONS ON CIRCUITS AND SYSTEMS FOR VIDEO TECHNOLOGY, VOL. xx, NO. x, xxx 20xx}%
{Shell \MakeLowercase{\underline{et al.}}: Bare Demo of IEEEtran.cls for IEEE Journals}

\maketitle

\makeatletter
\def\ps@IEEEtitlepagestyle{%
  \def\@oddfoot{\mycopyrightnotice}%
  \def\@oddhead{\hbox{}\@IEEEheaderstyle\leftmark\hfil\thepage}\relax
  \def\@evenhead{\@IEEEheaderstyle\thepage\hfil\leftmark\hbox{}}\relax
  \def\@evenfoot{}%
}
\def\mycopyrightnotice{%
  \begin{minipage}{\textwidth}
  \centering \scriptsize
  Copyright © 20xx IEEE. Personal use of this material is permitted. However, permission to use this material for any other purposes must be obtained from the IEEE by sending an email to pubs-permissions@ieee.org.
  \end{minipage}
}
\makeatother

\begin{abstract}
Existing deep embedding clustering methods fail to sufficiently utilize the available off-the-shelf information from feature embeddings and cluster assignments, limiting their performance. To this end, we propose a novel method, namely deep attention-guided graph clustering with dual self-supervision (DAGC). 
Specifically, DAGC first utilizes a heterogeneity-wise fusion module to adaptively integrate the features of the auto-encoder and the graph convolutional network in each layer and then uses a scale-wise fusion module to dynamically concatenate the multi-scale features in different layers. Such modules are capable of learning an informative feature embedding via an attention-based mechanism. 
In addition, we design a distribution-wise fusion module that leverages cluster assignments to acquire clustering results directly. 
To better explore the off-the-shelf information from the cluster assignments, we develop a dual self-supervision solution consisting of a soft self-supervision strategy with a Kullback-Leibler divergence loss and a hard self-supervision strategy with a pseudo supervision loss. 
Extensive experiments on nine benchmark datasets validate that our method consistently outperforms state-of-the-art methods. Especially, our method improves the ARI by more than 10.29\% over the best baseline. The code will be publicly available at \url{https://github.com/ZhihaoPENG-CityU/DAGC}.
\end{abstract}


\IEEEpeerreviewmaketitle

\section{Introduction}
\IEEEPARstart{C}{lustering} is one of the fundamental tasks in data analysis, which aims to categorize samples into multiple groups according to their intrinsic similarities, and has been successfully applied to many real-world applications such as image processing \cite{vidal2005generalized,wang2020deep,dang2020multi,wang2021decorrelated,jia2021clustering}, face recognition \cite{yang2020orthogonality,jia2020constrained,wu2020parallel}, and object detection \cite{peng2019active,jia2021multi,peng2021maximum}. 
Recently, with the booming of deep learning, numerous researchers have paid attention to deep embedding clustering analysis, which could effectively learn a clustering-friendly representation by extracting intrinsic patterns from the latent embedding space. 
For example, 
Hinton and Salakhutdinov \cite{hinton2006reducing} developed a deep auto-encoder (DAE) framework that first conducts embedding learning and then performs K-means \cite{macqueen1967some} to obtain clustering results. 
Xie \textit{et al.} \cite{xie2016unsupervised} designed a deep embedding clustering method (DEC) to perform embedding learning and cluster assignment jointly. 
Guo \textit{et al.} \cite{guo2017improved} improved DEC by introducing a reconstruction loss to preserve data structure. Although these DAE-based approaches obtain impressive improvement, they neglect the underlying topological structure among data, which has demonstrated its importance in various works \cite{wu2019simplifying,kim2021find,GNNBook2022}.

Recently, a series of works have been proposed to use graph convolutional networks (GCNs) \cite{kipf2016semi} to exploit the topological structure information. 
For instance, 
Kipf \textit{et al.} \cite{kipf2016variational} incorporated GCN into DAE and variational DAE, and proposed graph auto-encoder (GAE) and variational graph auto-encoder (VGAE), respectively. 
Pan \textit{et al.} \cite{pan2019learning} designed an adversarially regularized graph auto-encoder network (ARGA) to promote GAE. 
Wang \textit{et al.} \cite{wang2019attributed} incorporated graph attention networks \cite{velivckovic2018graph} into GAE for attributed graph clustering. 
Bo \textit{et al.} \cite{bo2020structural} fused GCN into DEC to consider the node content and topological structure information at the same time. 

\textcolor{black}{However, these works still suffer from the following drawbacks. 
First, they equate the importance of the features extracted from DAE and GCN, e.g., in \cite{bo2020structural}, the DAE and GCN features of a typical layer are averaged. Such a simple fusion strategy is not a good choice since those features contain different characteristic information.
Second, they neglect the multi-scale information embedded in different layers, which may lead to inferior clustering results. 
Third, they output two probability distributions capable of obtaining the final clustering results; however, the complex real-world datasets are usually agnostic and vastly different, so it is difficult to decide which one should be used to get the final clustering results. To the best of our knowledge, this is a decision-making dilemma for those kinds of deep graph clustering methods.
Last but not least, the previous approaches fail to adequately exploit the available information from the high-confidence clustering assignments.}

\begin{figure*}
	\centering
	\includegraphics [width=1.98\columnwidth]{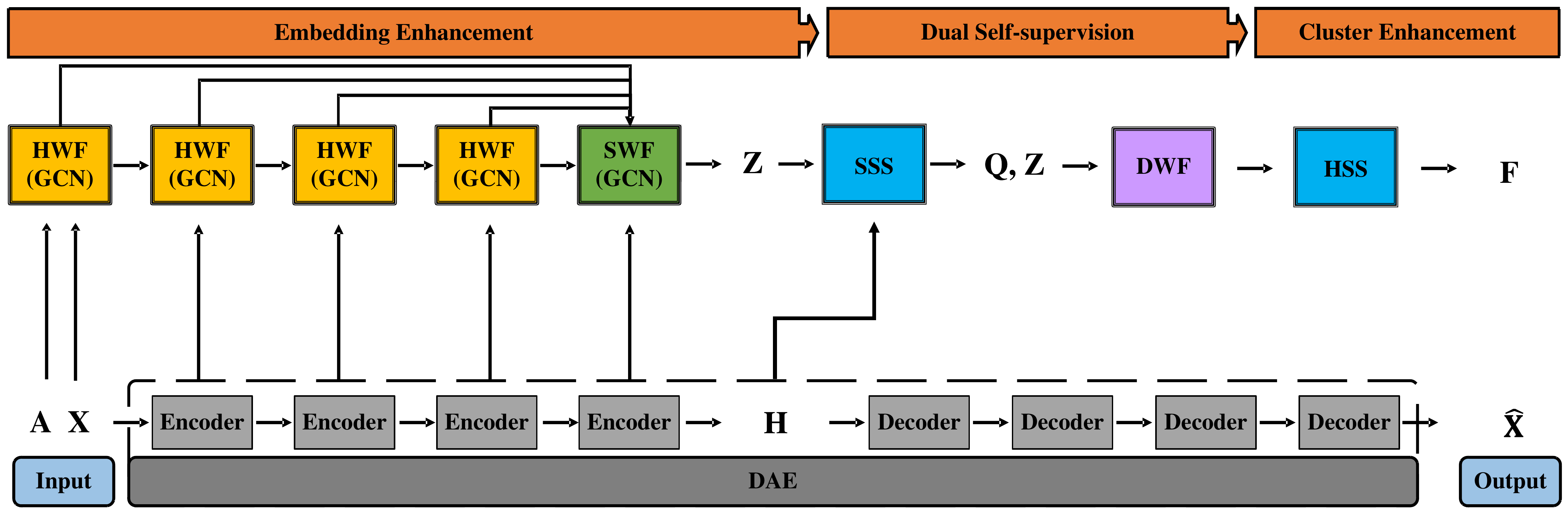}
    \setlength{\abovecaptionskip}{2pt}
    \setlength{\abovecaptionskip}{8pt}
	\caption{\textcolor{black}{The overall flowchart of the proposed method, namely deep attention-guided graph clustering with dual self-supervision (DAGC).} It consists of a DAE module, a heterogeneity-wise fusion (HWF) module, a scale-wise fusion (SWF) module, a distribution-wise fusion (DWF) module, a soft self-supervision (SSS) strategy, and a hard self-supervision (HSS) strategy. Specifically, HWF and SWF conduct the weight fusion in the sum and concatenation manner, respectively, where both modules involve a multilayer perceptron module, a normalization operation, and a GCN module. DWF uses a softmax function to infer a probability distribution. To achieve an end-to-end self-supervision, SSS drives the soft assignments to achieve distributions alignment between $\mathbf{Q}$ and $\mathbf{Z}$ distributions, and HSS transfers the cluster assignment to a hard one-hot encoding. The detailed architectures of HWF, SWF, DWF, and SSS are given in Figures \ref{fig: HWF-S}, \ref{fig: DWF}, and \ref{fig: DWFSS}, respectively.}
	\vspace{0.3cm}
	\label{fig: Our_framework}
\end{figure*}

% Our contribution
To address the above-mentioned drawbacks, we propose a novel deep embedding clustering method, focusing on exploiting the available off-the-shelf information from feature embeddings and cluster assignments. As shown in Figure \ref{fig: Our_framework}, the proposed method consists of a heterogeneity\footnote{Here, `heterogeneity' indicates the discrimination of feature structure, e.g., the DAE-based feature structure and the GCN-based feature structure.}-wise fusion (HWF) module, a scale-wise fusion (SWF) module, a distribution-wise fusion (DWF) module, a soft self-supervision (SSS) strategy, and a hard self-supervision (HSS) strategy. \textcolor{black}{A preliminary version of this work was published in ACM Multimedia 2021 \cite{peng2021attention}, which can be regarded as a special case of the current version that focuses on embedding enhancement via the HWF and SWF modules. However, the conference paper suffers from the decision-making dilemma concerning two learned probability distributions from DAE and GCN, i.e., which one should be selected as the final clustering assignment result. In summary, the main contributions of this journal paper are as follows:} 
\begin{itemize}
    % \item We first design the HWF module to adaptively integrate the DAE and GCN features within each layer. Then, we develop the SWF module to dynamically concatenate the multi-scale features from different layers. Such modules obey an attention-based mechanism to consider the off-the-shelf information to learn a informative feature embedding.
    \item \textcolor{black}{To handle the decision-making dilemma, we propose a \textbf{learning-aware fusion} module to adaptively fuse the learned data probability distributions to predict the clustering results.}
    \item \textcolor{black}{In addition, for the two learned distributions, we improve the soft self-supervision strategy to better preserve the distribution consistency alignment.}
    \item \textcolor{black}{Moreover, for the aforementioned fused distribution, we develop a hard self-supervision strategy with a pseudo supervision loss to employ the high-confidence clustering assignments to improve clustering performance.}
    \item \textcolor{black}{Extensive experiments on nine benchmark datasets validate that our method consistently outperforms the conference version \cite{peng2021attention}. For instance, on DBLP, our method improves the ARI by more than 14.80\%. In addition, the ablation studies and visualizations quantitatively and qualitatively validate the effectiveness of this method.}
\end{itemize}

We organize the rest of this paper as follows. Section \ref{sec: rw} briefly reviews the related works. Section \ref{sec: pm} introduces the proposed network architecture, followed by the experimental results and analyses in Section \ref{sec: eprm}. Finally, we conclude this paper in Section \ref{sec: con}.

\textit{Notation:} Throughout the paper, scalars are denoted by italic lower case letters, vectors by bold lower case letters, matrices by bold upper case ones, and operators by calligraphy ones, respectively. Let $\mathbf{X}$ be the input data, $\mathcal{V}$ be the node set, $\mathcal{E}$ be the edge set, and $\mathcal{G}=(\mathcal{V},\mathcal{E},\mathbf{X})$ be the undirected graph. $\mathbf{A} \in\mathbb{R}^{\emph{n}\times \emph{n}}$ denotes the adjacency matrix, $\mathbf{D} \in\mathbb{R}^{\emph{n}\times \emph{n}}$ denotes the degree matrix, and $\mathbf{I}\in\mathbb{R}^{\emph{n}\times \emph{n}}$ denotes the identity matrix. We summarize the main notations in Table \ref{tab: notation}.

\begin{table}[ht]
    \caption{Main notations and descriptions.}
    \label{tab: notation}
    \centering
        \begin{tabular}{cl|l}
        \hline\hline
        \multicolumn{2}{c|}{Notations} & Descriptions \\ 
        \hline
        $\mathbf{X}$&$\in\mathbb{R}^{\emph{n}\times \emph{d}}$                        & The input matrix                                      \\
        $\hat{\mathbf{X}}$&$\in\mathbb{R}^{\emph{n}\times \emph{d}}$                  & The reconstructed matrix                              \\
        $\mathbf{A}$&$\in\mathbb{R}^{\emph{n}\times \emph{n}}$                        & The adjacency matrix                                  \\
        $\mathbf{D}$&$\in\mathbb{R}^{\emph{n}\times \emph{n}}$                        & The degree matrix                                     \\
        $\mathbf{Z}_\emph{i}$&$\in\mathbb{R}^{\emph{n}\times {\emph{d}_\emph{i}}}$               & The GCN feature from the $i_{th}$ layer               \\
        $\mathbf{H}_\emph{i}$&$\in\mathbb{R}^{\emph{n}\times {\emph{d}_\emph{i}}}$               & The encoder feature from the $i_{th}$ layer           \\
        $\mathbf{M}_\emph{i}$&$\in\mathbb{R}^{\emph{n}\times 2}$               & The HWF weight matrix                               \\
        $\mathbf{Z}_\emph{i}^{'}$&$\in\mathbb{R}^{\emph{n}\times {\emph{d}_\emph{i}}}$           & The HWF combined feature                            \\
        $\mathbf{U}$&$\in\mathbb{R}^{\emph{n}\times \left(\emph{l}+1\right)}$                        & The SWF weight matrix                               \\
        $\mathbf{H}$&$\in\mathbb{R}^{\emph{n}\times {\emph{d}_\emph{l}}}$                        & The DAE extracted feature                             \\
        $\mathbf{Q}$&$\in\mathbb{R}^{\emph{n}\times \emph{k}}$                        & The distribution obtained from DAE                     \\
        $\mathbf{Z}$&$\in\mathbb{R}^{\emph{n}\times \emph{k}}$                        & The distribution obtained from SWF                   \\
        $\mathbf{P}$&$\in\mathbb{R}^{\emph{n}\times \emph{k}}$                        & The auxiliary distribution                     \\
        $\mathbf{V}$&$\in\mathbb{R}^{\emph{n}\times 2}$                        & The DWF weight matrix                               \\
        $\mathbf{F}$&$\in\mathbb{R}^{\emph{n}\times \emph{k}}$                        & The DWF combined feature                            \\
        \hline
        $\emph{n}$ &                         & The number of samples                                 \\
        $\emph{d}$ &                         & The dimension of $\mathbf{X}$                         \\
        $\emph{d}_\emph{i}$ &                & The dimension of the $i_{th}$ latent feature          \\
        $\emph{l}$    &                      & The number of network layers                          \\
        $\emph{k}$   &                       & The number of clusters                                \\
        $\hat{\emph{k}}$  &                  & The number of neighbors for KNN graph                 \\
        $\emph{r}$    &                      & The threshold value for pseudo supervision            \\
        \hline
        $\cdot \| \cdot $  &                 & The concatenation operation                           \\
        % $\left\|\cdot\right\|_F$  &          & The Frobenius norm                                    \\
        \hline\hline   
        \end{tabular}
\end{table}

\begin{figure*}[]
	\centering
	\includegraphics [width=1.68\columnwidth]{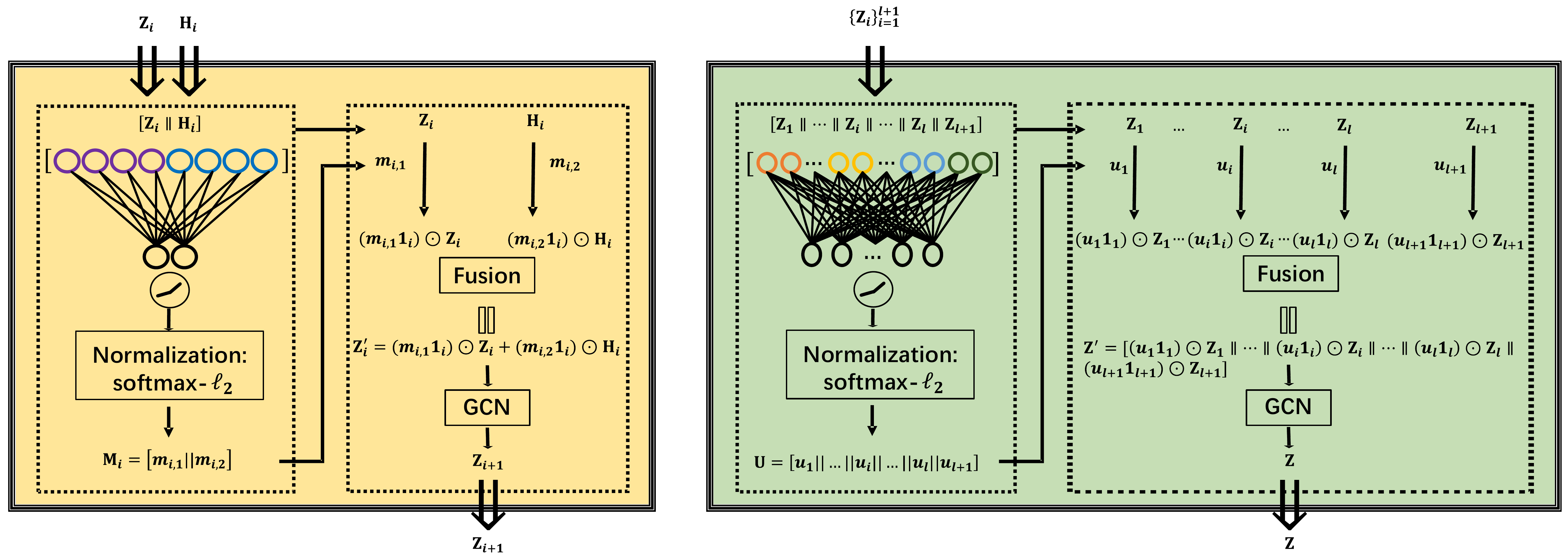}
    \caption{Illustration of the architectures of the HWF module (\textbf{left}) and the SWF module (\textbf{right}). The HWF module fuses the GCN feature $\mathbf{Z}_{i}$ and the DAE feature $\mathbf{h}_\emph{i}$ to obtain $\mathbf{Z}_{i+1}$ via a weighted sum form, while the SWF module combines the multi-scale weighted features in a feature concatenation manner. More specifically, we first learn the weights through the attention-based mechanism (the left dashed box in the triple-solid line box) and then integrate the corresponding features through the weighted fusion (the right dashed box in the triple-solid line box). Here, $\Downarrow$ represents the input and output actions.}
	\label{fig: HWF-S}
\end{figure*}

\section{Related Work}\label{sec: rw}

DAE is a typical deep neural network that allows computational models composed of multiple processing layers to learn data representations with multiple levels of abstraction. Recently, benefiting from the powerful representation ability of DAE, deep embedding clustering has achieved remarkable development \cite{dong2020unsupervised,huang2020deep,wang2020cluster,han2019learning,ou2020multimodal,wang2021dynamic,DCRN}. 
For example, Hinton and Salakhutdinov \cite{hinton2006reducing} used DAE to extract the feature representation of input data, on which K-means \cite{macqueen1967some} is performed to obtain the clustering results. 
\cite{xie2016unsupervised} jointly conducted embedding learning and cluster assignment in an iterative optimization manner. 
The improved DEC (IDEC) \cite{guo2017improved} enhanced the clustering performance by adding a reconstruction loss function into DEC. 
\textcolor{black}{A series of works \cite{wang2020deep,dang2020multi,hu2021akm} introduced multi-view information into the DAE framework to further improve embedding learning.}
However, these DAE-based methods neglect the underlying topological structure among data, which has demonstrated its effectiveness for data clustering \cite{wu2019simplifying,kim2021find,zhang2022rethinking,KDD22_GraphAKD,GNNBook2022}, thus limiting their performance.

Graph embedding is a new paradigm for clustering to capture the topological structure among samples \cite{wang2020gcn,zhang2020deep,chen2021multi,kim2021find,he2021consistency,wang2022duality,hu2022class}, and 
many recent approaches \cite{markovitz2020graph,park2019symmetric,goyal2018graph,li2019deepgcns,huang2021combining} have explored GCN to achieve graph embedding. 
For instance, 
Kipf and Welling \cite{kipf2016variational} provided GAE and VGAE methods by incorporating GCN into DAE and variational DAE frameworks, respectively. 
\cite{pan2019learning} extended GAE by introducing a designed adversarial regularization. 
Wang \textit{et al}. \cite{wang2019attributed} merged GAE and the graph attention network \cite{velivckovic2018graph} to build a deep attentional embedding framework for attributed graph clustering (DAEGC). 
The structural deep clustering network (SDCN) \cite{bo2020structural} fused the node content and topological structure information to achieve deep embedding clustering. 
Peng \textit{et al}. \cite{peng2021attention} designed the attention-driven graph clustering network (AGCN) to merge numerous features to enhance the embedding learning via an adaptive mechanism. 
The deep fusion clustering network (DFCN) \cite{tu2021deep} exploited a designed fusion strategy to combine the DAE and GAE frameworks to merge the node attribute and topological structure information. 
\textcolor{black}{He \textit{et al}. \cite{he2022parallelly} developed an adaptive graph convolutional clustering (AGCC) model to update the graph structure and the data representation layer by layer.}

\textcolor{black}{Optimizing a deep clustering network is a fundamental yet challenging task because there are no ground-truth labels as supervision. Previous works \cite{xie2016unsupervised,guo2017improved,wang2019attributed,bo2020structural,peng2021attention} minimize the Kullback-Leibler (KL) divergence to tackle this challenge, and its effectiveness has been proven.} Specifically, it first uses the Student’s t-distribution \cite{helmert1876genauigkeit,student1908probable} as a kernel function to measure the similarity between the extracted feature $\mathbf{h}_\emph{i}$ and its corresponding centroid vector $\boldsymbol{\mu}_\emph{j}$, in which the measured similarity can be regarded as a probability distribution $\mathbf{Q}$ with its $\emph{i}, \emph{j}$-th element being 
\begin{equation}
\begin{aligned}
&\emph{q}_{\emph{i},\emph{j}} = \frac{(1+\|\mathbf{h}_\emph{i}-\boldsymbol{\mu}_{\emph{j}}\|^2/\alpha)^{-\frac{\alpha+1}{2}}}{ \sum_{\emph{j}^{'}} (1+\|\mathbf{h}_\emph{i}-\boldsymbol{\mu}_{\emph{j}^{'}}\|^2/\alpha)^{-\frac{\alpha+1}{2}} },
\label{eq: KL-DEC-q}
\end{aligned}
\end{equation}
where $\alpha$ is set to $1$. Then, it implements the KL divergence minimization between $\mathbf{Q}$ and a distribution $\mathbf{B}$ with the $\emph{i}, \emph{j}$-th element $b_{i,\emph{j}}=\frac{ q_{i,\emph{j}}^{2}/\sum_{i} q_{i,\emph{j}} }{\sum_{\emph{j}}^{'} q_{i,\emph{j}^{'}}^{2}/\sum_{i} q_{i,\emph{j}^{'}}}$, i.e., 
\begin{equation}
\begin{aligned}
\min KL(\mathbf{B}, \mathbf{Q}) =\sum_i\sum_j{b_{i,\emph{j}} log{\frac{b_{i,\emph{j}}}{q_{i,\emph{j}}}}},
\label{eq: KL-DEC}
\end{aligned}
\end{equation}
where $KL\left(\cdot,\cdot\right)$ is the Kullback-Leibler divergence function that measures the distance between two distributions. Such an auxiliary distribution generation normalizes the high-confidence probability value as a large value, capable of learning the high-confidence assignments. However, previous works fail to sufficiently utilize the available off-the-shelf information from high-confidence clustering assignments, inevitably leading to inferior clustering results.

\section{Proposed Method}\label{sec: pm}
\textcolor{black}{Figure \ref{fig: Our_framework} illustrates the architecture of the proposed method, where we will detail the main components in what follows.}

\subsection{Heterogeneity-Wise Fusion}

We first exploit a DAE module with a series of encoders and decoders to extract the latent representation by adapting the reconstruction loss, i.e.,
\begin{equation}
\begin{aligned}
&\mathcal{L}_{R} = \left\| \mathbf{X} - \hat{\mathbf{X}} \right\|^2_F,
\label{eq: AE}
\end{aligned}
\end{equation}
where $\mathbf{X}$ and $\hat{\mathbf{X}}$ denote the input matrix and the reconstructed matrix, respectively. Here, $\mathbf{H}_{\emph{0}}=\mathbf{X}$, $\hat{\mathbf{H}}_{\emph{l}}=\hat{\mathbf{X}}$, $\mathbf{H}_{\emph{i}}= \phi ( \mathbf{W}_{\emph{i}}^{e}\mathbf{H}_{\emph{i}-1}+ \mathbf{b}_{\emph{i}}^{e})$, $\hat{\mathbf{H}}_{\emph{i}}= \phi ( \mathbf{W}_{\emph{i}}^{d}\hat{\mathbf{H}}_{\emph{i}-1}+ \mathbf{b}_{\emph{i}}^{d})$, where $\mathbf{H}_{\emph{i}}$ and $\hat{\mathbf{H}}_{\emph{i}}$ denote the encoder and decoder outputs from the $i_{th}$ layer, respectively, $\emph{l}$ denotes the number of encoder/decoder layers, $\mathbf{W}_{\emph{i}}^{e}$, $\mathbf{b}_{i}^{e}$, $\mathbf{W}_{\emph{i}}^{d}$, and $\mathbf{b}_{i}^{d}$ denote the network weight and bias of the $i_{th}$ encoder and decoder layer, respectively, and $\phi ( \cdot )$ denotes an activation function, such as Tanh or ReLU \cite{glorot2011deep}. Particularly, we set $\mathbf{H}=\mathbf{H}_\emph{l}$ for convenience. In addition, we denote the GCN feature learned from the $\emph{i}_{th}$ layer as $\mathbf{Z}_\emph{i}\in\mathbb{R}^{\emph{n}\times \emph{d}_\emph{i}}$ with $\emph{d}_{i}$ being the dimension of the $\emph{i}_{th}$ layer, where $\mathbf{Z}_{0}=\mathbf{X}$. 
\textcolor{black}{
Previous works (e.g., SDCN \cite{bo2020structural}) combine the heterogeneity-wise representation on the $i_{th}$ layer ($\mathbf{Z}_i$ and $\mathbf{H}_i$) via a fixed fusion strategy (i.e., $\mathbf{Z}_{i}^{'}=0.5 \mathbf{Z}_i+0.5\mathbf{H}_i$) to enhance representation learning. However, such a fusion strategy is simple but unreasonable since the heterogeneity-wise representations $\mathbf{Z}_i$ and $\mathbf{H}_i$ owe different characteristic information. To this end, we propose a learning-aware fusion strategy to develop an adaptive fusion strategy to dynamically weight $\mathbf{Z}_i$ and $\mathbf{H}_i$.} Specifically, to learn the corresponding attention coefficients of $\mathbf{Z}_\emph{i}$ and $\mathbf{H}_\emph{i}$, we first concatenate them as $[\mathbf{Z}_\emph{i}\|\mathbf{H}_\emph{i}]\in\mathbb{R}^{\emph{n}\times {\emph{2d}_\emph{i}}}$ and then build a fully connected layer parametrized by a weight matrix $\mathbf{W}_{\emph{i}}^{a}\in\mathbb{R}^{{\emph{2d}_\emph{i}}\times 2}$. Afterwards, we apply the LeakyReLU (LReLU) \cite{maas2013rectifier} on the product between $\left[\mathbf{Z}_\emph{i}\|\mathbf{H}_\emph{i}\right]$ and $\mathbf{W}_{\emph{i}}^{a}$, and normalize the output of the LReLU unit via the softmax function and the $\ell_2$ normalization (i.e., `softmax-$\ell_{2}$' normalization). Formally, we formulate the prediction of the corresponding attention coefficients as
\begin{equation}
\begin{aligned}
\mathbf{M}_\emph{i}=[\mathbf{m}_{i,1}\| \mathbf{m}_{i,2}]=
\Upsilon_{\textit{A}}(\left[\mathbf{Z}_\emph{i}\|\mathbf{H}_\emph{i}\right]\mathbf{W}_{\emph{i}}^{a}),
\label{eq: HWF-A}
\end{aligned}
\end{equation}
where $\mathbf{M}_\emph{i}\in\mathbb{R}^{\emph{n}\times 2}$ is the attention coefficient matrix with entries being greater than $0$, $\mathbf{m}_{i,1}$ and $\mathbf{m}_{i,2}$ are the weight vectors for measuring the importance of $\mathbf{Z}_\emph{i}$ and $\mathbf{H}_\emph{i}$, respectively, and 
$\Upsilon_{\textit{A}}(\cdot)=\ell_{2}\left( softmax \left(LReLU\left(\cdot\right)\right) \right)$. 
Thus, we can adaptively fuse the GCN feature $\mathbf{Z}_\emph{i}$ and the DAE feature $\mathbf{H}_\emph{i}$ on the $\emph{i}_{th}$ layer as
\begin{equation}
\begin{aligned}
& \mathbf{Z}_\emph{i}^{'}= \left( \mathbf{m}_{i,1}\mathbf{1}_\emph{i} \right) \odot \mathbf{Z}_\emph{i} + \left( \mathbf{m}_{i,2}\mathbf{1}_\emph{i} \right) \odot \mathbf{H}_\emph{i},
\label{eq: HWF}
\end{aligned}
\end{equation}
where $\mathbf{1}_\emph{i}\in\mathbb{R}^{1\times \emph{d}_\emph{i}}$ denotes the vector of all ones, and $\odot$ denotes the Hadamard product of matrices. Then, we use the resulting matrix $\mathbf{Z}_\emph{i}^{'}\in\mathbb{R}^{\emph{n}\times \emph{d}_\emph{i}}$ as the input of the $(\emph{i}+1)_{th}$ GCN layer to learn the representation $\mathbf{Z}_{\emph{i}+1}$, i.e.,
\begin{equation}
\begin{aligned}
& \mathbf{Z}_{\emph{i}+1}= LReLU(\mathbf{D}^{-\frac{1}{2}}(\mathbf{A}+\mathbf{I})\mathbf{D}^{-\frac{1}{2}}\mathbf{Z}_\emph{i}^{'}\mathbf{W}_\emph{i}),
\label{eq: HWF-output}
\end{aligned}
\end{equation}
where $\mathbf{W}_\emph{i}$ denotes the weight matrix from the $i_{th}$ GCN layer, and $\mathbf{D}^{-\frac{1}{2}}(\mathbf{A}+\mathbf{I})\mathbf{D}^{-\frac{1}{2}}$ normalizes $\mathbf{A}$ by using renormalization with a self-loop normalized $\mathbf{A}$ and the corresponding $\mathbf{D}$. 

\subsection{Scale-Wise Fusion}
As aforementioned, previous works neglect the off-the-shelf multi-scale information embedded in different layers, which is of great importance for embedding learning. To this end, we propose the SWF module to concatenate the multi-scale features from different layers via an attention-based mechanism. The right border of Figure \ref{fig: HWF-S} shows the overall architecture of SWF.

We aggregate the multi-scale features with a concatenation manner to dynamically combine various scale features with different dimensions. Afterwards, we build a fully connected layer parametrized by a weight matrix $\mathbf{W}^\emph{s}\in\mathbb{R}^{(\sum_{j=1}^{l}\emph{d}_j+\emph{d}_\emph{l}) \times (\emph{l}+1)}$ to capture the relationship among the multi-scale features.
Formally, we formulate the whole process as 
\begin{equation}
\begin{aligned}
\mathbf{U}=\Upsilon_{\textit{A}}\left( \Xi_{j=1}^{\emph{l}+1}\mathbf{Z}_j\mathbf{W}^{s}\right),
\label{eq: SWF-A}
\end{aligned}
\end{equation}
where $\Xi_{j=1}^{\emph{l}+1}\mathbf{Z}_j=\left[\mathbf{Z}_1\|\cdots\|\mathbf{Z}_{\emph{l}}\|\mathbf{Z}_{\emph{l}+1}\right]$ denotes the concatenation operation of multiple elements. We then conduct the feature fusion as
\begin{equation}
\begin{aligned}
\mathbf{Z}^{'} = \Xi_{j=1}^{\emph{l}+1}\left(\left(\mathbf{u}_{j}\mathbf{1}_{j}\right)\odot\mathbf{Z}_{\emph{j}}\right),
\label{eq: SWF}
\end{aligned}
\end{equation}
where $\mathbf{u}_{j}$ is the $\emph{j}$-th element of $\mathbf{U}$, i.e., $\Xi_{j=1}^{\emph{l}+1}\mathbf{u}_j=\mathbf{U}$. In addition, we use a Laplacian smoothing operator \cite{li2018deeper} and the softmax function to make the fused feature $\mathbf{Z}^{'}$ as a reasonable probability distribution, i.e.,
\begin{equation}
\begin{aligned}
& \mathbf{Z}= softmax(\mathbf{D}^{-\frac{1}{2}}(\mathbf{A}+\mathbf{I})\mathbf{D}^{-\frac{1}{2}}\mathbf{Z}^{'}\mathbf{W}), 
\label{eq: Our-output}
\end{aligned}
\end{equation}
where $\mathbf{W}$ denotes the learnable parameters. 

\begin{figure}[]
	\centering
	\includegraphics [width=0.88\columnwidth]{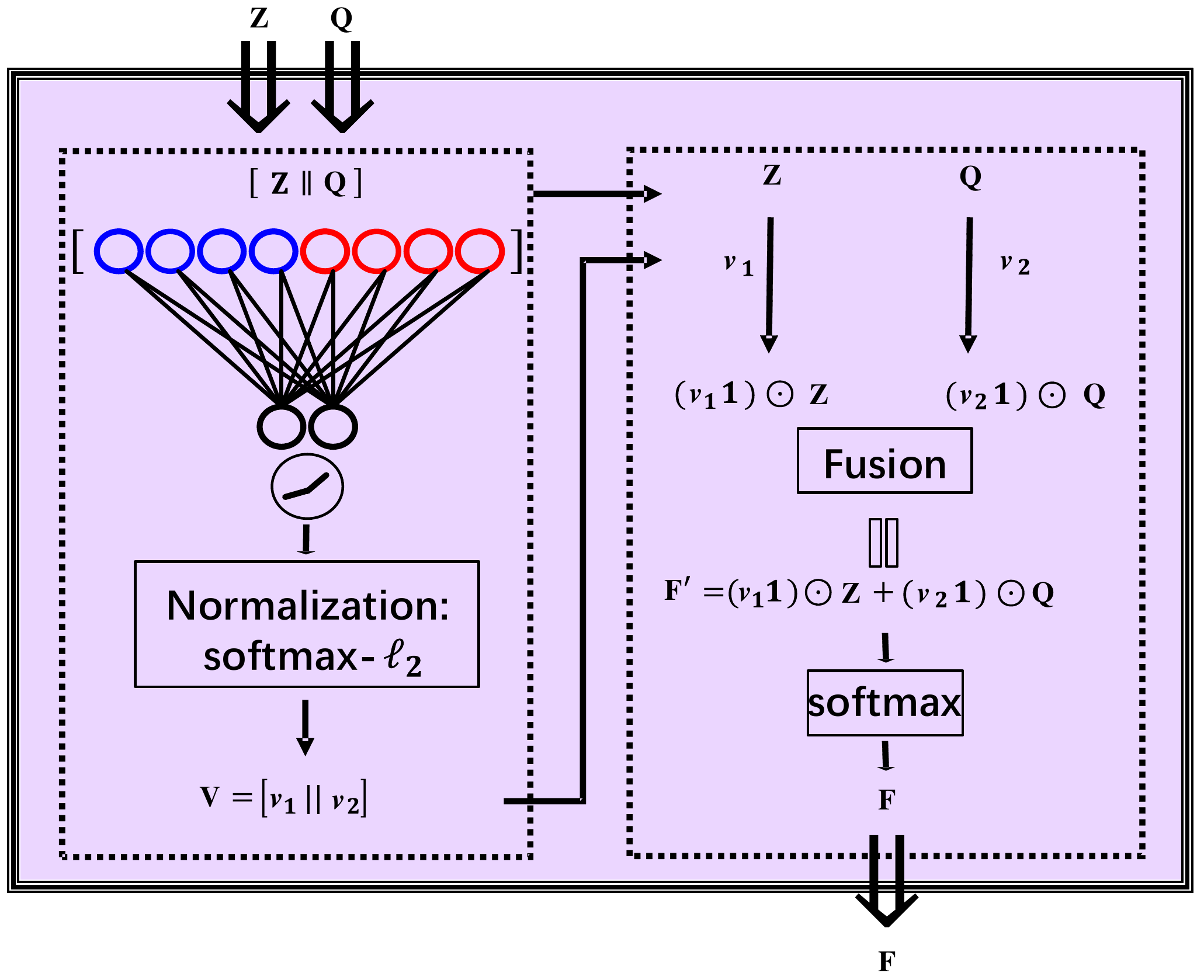}
    \caption{The architecture of DWF module. The DWF module dynamically combines the distributions $\mathbf{Z}$ and $\mathbf{Q}$ to learn the final probability distribution, where we can directly obtain the predicted cluster label.}
	\label{fig: DWF}
\end{figure}

\subsection{Distribution-Wise Fusion}
\textcolor{black}{
As we obtain the feature $\mathbf{H}$ from DAE, we can exploit it to calculate its cluster center embedding $\boldsymbol{\mu}$ with K-means. Afterward, we measure the similarity between the extracted feature $\mathbf{h}_\emph{i}$ and its corresponding centroid vector $\boldsymbol{\mu}_\emph{j}$, in which the measured similarity can be regarded as a probability distribution $\mathbf{Q}$ with its $\emph{i}, \emph{j}$-th element being $\emph{q}_{\emph{i},\emph{j}} = \frac{(1+\|\mathbf{h}_\emph{i}-\boldsymbol{\mu}_{\emph{j}}\|^2)^{-1}}{ \sum_{\emph{j}^{'}} (1+\|\mathbf{h}_\emph{i}-\boldsymbol{\mu}_{\emph{j}^{'}}\|^2)^{-1} }$, following Eq. (\ref{eq: KL-DEC-q}).}
Both $\mathbf{Z}$ and $\mathbf{Q}$ can generate the final clustering results; however, it is challenging to choose which one to obtain the final clustering result within different scenarios. To the best of our knowledge, this is an unsolved decision-making dilemma commonly existing in the previous deep graph clustering methods. To handle this challenge, we propose the DWF module to fuse the learned data probability distributions in an attention-driven manner to predict cluster labels. Figure \ref{fig: DWF} shows the overall architecture.

Specifically, we first learn the importance of $\mathbf{Z}$ and $\mathbf{Q}$ by an attention-based mechanism, i.e.,
\begin{equation}
\begin{aligned}
\mathbf{V}=[\mathbf{\mathbf{v}}_{1}\| \mathbf{\mathbf{v}}_{2}]=\Upsilon_{\textit{A}}\left(\left[\mathbf{Z}\|\mathbf{Q}\right]\mathbf{\hat{W}}\right),
\label{eq: DWF-A}
\end{aligned}
\end{equation}
where $\mathbf{V}\in\mathbb{R}^{\emph{n}\times 2}$ is the attention coefficient matrix, $\mathbf{\hat{W}}$ is a learned weight matrix via a fully connected layer. We then adaptively leverage $\mathbf{Z}$ and $\mathbf{Q}$ as
\begin{equation}
\begin{aligned}
& \mathbf{F} = \left( \mathbf{\mathbf{v}}_{1}\mathbf{1} \right) \odot \mathbf{Z} + \left( \mathbf{\mathbf{v}}_{2}\mathbf{1} \right) \odot \mathbf{Q},
\label{eq: DWF-1}
\end{aligned}
\end{equation}
where $\mathbf{1}\in\mathbb{R}^{1\times k}$ denotes the vector of all ones. Finally, we apply the softmax function to normalize $\mathbf{F}$ with
\begin{equation}
\begin{aligned}
& \mathbf{F} = softmax\left( \mathbf{F} \right) \quad \rm{s.t.} \quad \sum_{\emph{j}=1}^{\emph{k}}\emph{f}_{\emph{i},\emph{j}}=1, \quad \emph{f}_{\emph{i},\emph{j}}>0,
\label{eq: DWF}
\end{aligned}
\end{equation}
where $\emph{f}_{i,\emph{j}}$ is the element of $\mathbf{F}$. When the network is well-trained, we can directly infer the predicted cluster label through $\mathbf{F}$, i.e.,
\begin{equation}
\begin{aligned}
& y_{i}=\mathop{\arg\max}_{\emph{j}} \emph{f}_{i,\emph{j}} \quad \rm{s.t.} \quad \emph{j}=1,\cdots,\emph{k},
\label{eq: clustering_result}
\end{aligned}
\end{equation}
where $y_{i}$ is the predicted label of $\mathbf{x}_{i}$. In this way, the cluster structure can be represented explicitly in $\mathbf{F}$. 

\subsection{Dual Self-supervision}
As unsupervised clustering lacks reliable guidance, we propose a novel dual self-supervision scheme that combines a soft self-supervision strategy with a Kullback-Leibler (KL) divergence loss and a hard self-supervision strategy with a pseudo supervision loss to guide the overall network training, as illustrated in Figure \ref{fig: DWFSS}.

\subsubsection{Soft Self-supervision}
\label{subsec: SSS}
Since we take advantage of the high-confidence assignments to iteratively refine the clusters by utilizing the soft assignments (i.e., the probability distributions $\mathbf{Q}$ and $\mathbf{Z}$), we term this supervision strategy as the \textbf{soft self-supervision} strategy. Concretely, since $\mathbf{Z}$ involves the graph information through the HWF and SWF modules, we first derive an auxiliary distribution $\mathbf{P}$ via $\mathbf{Z}$ by normalizing per cluster after squaring $\emph{z}_{\emph{i},\emph{j}}$, i.e., 
\begin{equation}
\begin{aligned}
&\emph{p}_{\emph{i},\emph{j}}=\frac{ \emph{z}_{\emph{i},\emph{j}}^{2}/\sum_{\emph{i}^{'}=1}^\emph{n}  \emph{z}_{\emph{i}^{'},\emph{j}} }{\sum_{\emph{j}^{'}=1}^\emph{k} \emph{z}_{\emph{i},\emph{j}^{'}}^{2}/\sum_{\emph{i}^{'}=1}^\emph{n}\sum_{\emph{j}^{'}=1}^\emph{k}  \emph{z}_{\emph{i}^{'},\emph{j}^{'}}},
\label{eq: KL-p}
\end{aligned}
\end{equation}
where $0\leq \emph{p}_{\emph{i},\emph{j}}\leq 1$ is the element of $\mathbf{P}$. Then, we minimize the KL divergence not only between a learned distribution and its auxiliary distribution (i.e., $KL\left(\mathbf{P}, \mathbf{Z}\right)$ and $KL\left(\mathbf{P}, \mathbf{Q}\right)$), but also between both two learned distributions (i.e., $KL\left(\mathbf{Z}, \mathbf{Q}\right)$) to promote a highly consistent distribution alignment to train our model, i.e.,
\begin{equation}
\begin{aligned}
\mathcal{L}_{S} 
&=\lambda_1\!*\!(KL\left(\mathbf{P}, \mathbf{Z}\right) \!+\! KL\left(\mathbf{P}, \mathbf{Q})\right) \!+\! \lambda_2*KL\left(\mathbf{Z}, \mathbf{Q}\right) \\
&=\lambda_1\sum_\emph{i}^\emph{n}\sum_\emph{j}^\emph{k}{ \emph{p}_{\emph{i},\emph{j}}log{\frac{\emph{p}_{\emph{i},\emph{j}}^2}{z_{i,\emph{j}} \emph{q}_{\emph{i},\emph{j}}}}} \!+\! \lambda_2\sum_\emph{i}^\emph{n}\sum_\emph{j}^\emph{k}{ z_{i,\emph{j}} log{\frac{z_{i,\emph{j}}}{\emph{q}_{\emph{i},\emph{j}}}}},
\label{eq: KL}
\end{aligned}
\end{equation}
where $\lambda_1>0$ and $\lambda_2>0$ are the trade-off parameters.

\subsubsection{Hard Self-supervision}
\begin{figure}[]
	\centering
	\includegraphics [width=0.88\columnwidth]{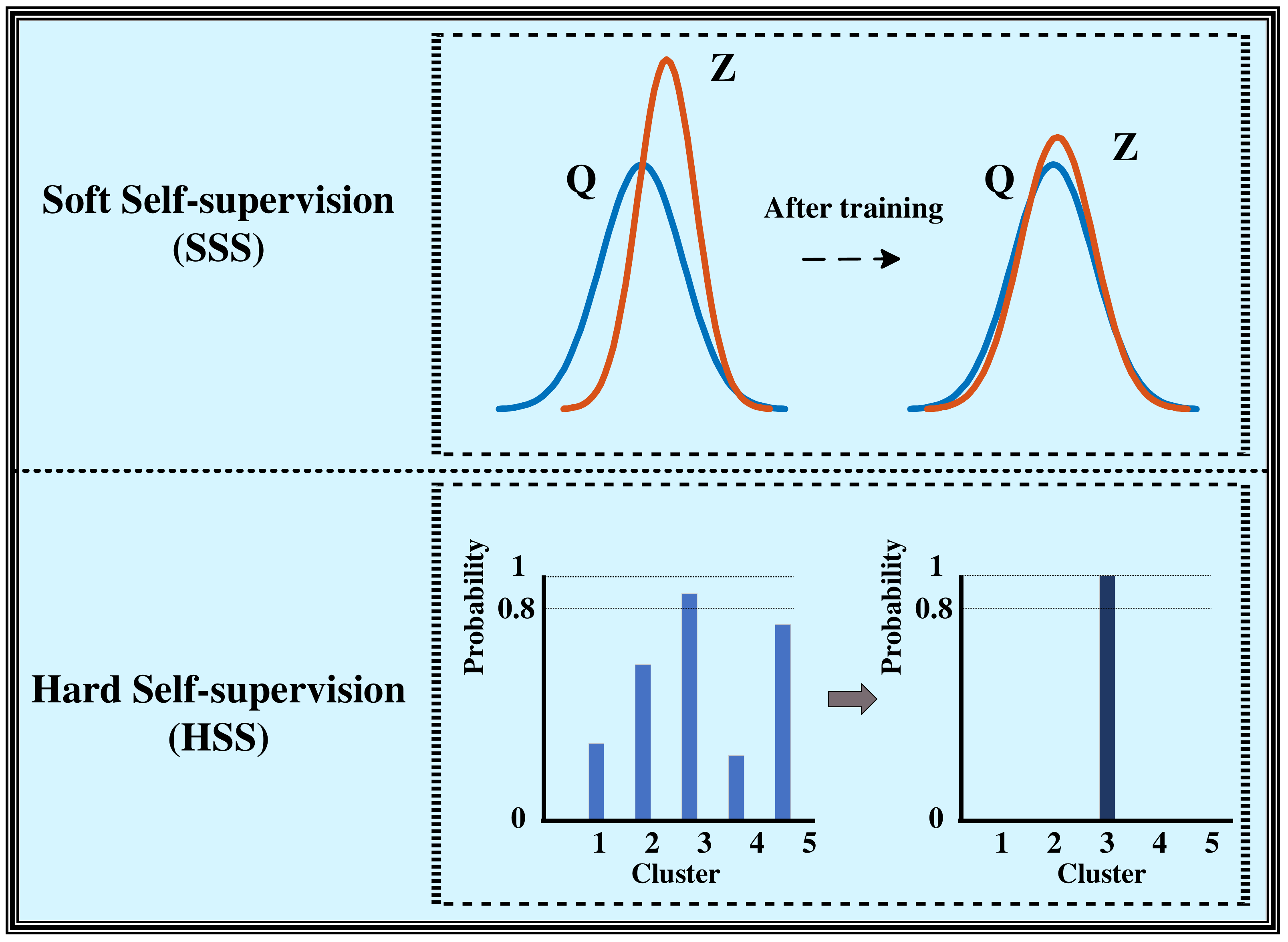}
    \caption{The illustration of the proposed dual self-supervision solution. It exploits a soft self-supervision strategy and a hard self-supervision strategy to effectively train the proposed network in an end-to-end manner. Such strategies iteratively refine the network training by learning from high-confidence assignments.}
	\label{fig: DWFSS}
\end{figure}

\textcolor{black}{
Although the soft self-supervision strategy has become a helpful tool for unsupervised clustering, it preserves the low-confidence predicted probabilities, limiting the clustering performance.} To further make use of the available off-the-shelf information from the cluster assignments, we introduce the pseudo supervision technique \cite{lee2013pseudo} and set the pseudo-label $\hat{y}_{i}$ as $\hat{y}_{i}={y}_{i}$. Considering that the pseudo-labels may contain many incorrect labels, we select the high-confidence ones as supervisory information by a large threshold $\emph{r}$, i.e.,
\begin{equation}
\emph{g}_{i,\emph{j}}=\left\{\begin{array}{ll}
1 & \text { if } f_{i,\emph{j}} \geq \text { \emph{r}, } \\
0 & \text { otherwise. }
\end{array}\right.
\end{equation}
In the experiment, we set $r=0.8$. Then, we leverage the high-confidence pseudo-labels to supervise the network training, i.e., 
\textcolor{black}{
\begin{equation}
\begin{aligned}
\mathcal{L}_{H} 
&= \lambda_3\sum_\emph{i}\sum_{\emph{j}} \emph{g}_{i,\emph{j}} * \Upsilon_{\textit{CE}}(\mathbf{\emph{f}}_{i,\emph{j}},\Upsilon_{\textit{OH}}(\hat{y}_{i})),
\label{eq: PL}
\end{aligned}
\end{equation}
where $\lambda_3>0$ is the trade-off parameter, $\Upsilon_{\textit{CE}}$ denotes the cross-entropy \cite{de2005tutorial} loss, and $\Upsilon_{\textit{OH}}$ transforms $\hat{\emph{y}}_\emph{i}$ to its one-hot form. }
As shown in Figure \ref{fig: DWFSS}, the pseudo-labels transfer the cluster assignment to the hard one-hot encoding, we thus name it as \textbf{hard self-supervision} strategy. 

\textcolor{black}{In addition, we have empirically observed that only using HSS does not perform well in all scenarios. The reason may be that the distribution probability values are small in some situations, making a weak self-supervision for guiding network training with the HSS strategy. To this end, we combine the SSS and HSS strategies together to drive the network training.} Combining Eqs. (\ref{eq: AE}), (\ref{eq: KL}), and (\ref{eq: PL}), our overall loss function can be written as
\begin{equation}
\begin{aligned}
&\mathcal{L} = \min_{\mathbf{F}}\left( \mathcal{L}_{R}+\mathcal{L}_{S}+\mathcal{L}_{H}\right).
\label{eq: Our_loss}
\end{aligned}
\end{equation}
The whole training process is shown in Algorithm \ref{alg1}. 

\begin{algorithm}
	\caption{Training process of our method}
	\label{alg1}
	\begin{algorithmic}
		\REQUIRE Input matrix $\mathbf{X}$; Adjacency matrix $\mathbf{A}$; Cluster number $\emph{k}$; Trade-off parameters $\lambda_1,\lambda_2,\lambda_3$; Maximum iterations $i_\emph{MaxIter}$;\\
		\ENSURE Reconstructed matrix $\hat{\mathbf{X}}$; Clustering result $\mathbf{y}$;\\
	\end{algorithmic}
	\begin{algorithmic}[1]
    	\STATE Initialization: $\emph{l}=4$, $i_\emph{Iter} = 1$; $\mathbf{Z}_0 = \mathbf{X}$; $\mathbf{H}_0 = \mathbf{X}$;
    	\STATE Initialize the parameters of the DAE network;
		\WHILE{$i_\emph{Iter} < i_\emph{MaxIter}$}
		\STATE Obtain the feature $\mathbf{H}$ by Eq. (\ref{eq: AE});
		\STATE Obtain the feature $\mathbf{Z}$ via Eq. (\ref{eq: Our-output});
		\STATE Obtain the cluster center embedding $\boldsymbol{\mu}$ with K-means based on the feature $\mathbf{H}$;
		\STATE Calculate the distribution $\mathbf{Q}$ via Eq. (\ref{eq: KL-DEC-q});
		\STATE Calculate the distribution $\mathbf{P}$ via Eq. (\ref{eq: KL-p});
		\STATE Calculate the distribution $\mathbf{F}$ via Eq. (\ref{eq: DWF});
		\STATE Conduct the soft self-supervision via Eq. (\ref{eq: KL});
		\STATE Conduct the hard self-supervision via Eq. (\ref{eq: PL});
		\STATE Minimize the overall loss function via Eq. (\ref{eq: Our_loss});
		\STATE Conduct the back propagation and update parameters in the proposed network;
		\STATE $i_\emph{Iter} = i_\emph{Iter} + 1$;
		\ENDWHILE \\
		\STATE Calculate the clustering results $\mathbf{y}$ with $\mathbf{F}$ by Eq. (\ref{eq: clustering_result});
	\end{algorithmic}
\end{algorithm}

\subsection{Computational Complexity Analysis}
For the DAE module, the time complexity is $\mathcal{O}(n\sum_{i=2}^{l}d_{i-1}d_{i})$. For the GCN module, as the operation can be computed efficiently using sparse matrix computation, the time complexity is only $\mathcal{O}(|\mathcal{E}|\sum_{i=2}^{l}d_{i-1}d_{i})$ according to \cite{pan2019learning}. For Eq. (\ref{eq: KL-DEC-q}), the time complexity is $\mathcal{O}(nk+n\log n)$ based on \cite{xie2016unsupervised}. For HWF, SWF, and DWF modules, the total time complexity is $\mathcal{O}(\sum_{i=1}^{l-1}(d_{i}))+\mathcal{O}((\sum_{i=1}^{l+1} d_{i})(l+1))+\mathcal{O}(k)$. Thus, the overall computational complexity of Algorithm \ref{alg1} in one iteration is about $\mathcal{O}(n\sum_{i=2}^{l}d_{i-1}d_{i} + |\mathcal{E}|\sum_{i=2}^{l}d_{i-1}d_{i} + (n+1)k+n\log n + \sum_{i=1}^{l-1}(d_{i}) + (\sum_{i=1}^{l+1} d_{i})(l+1))$.

\begin{table}[]
\centering
\caption{\textcolor{black}{Description of the adopted datasets.}}
\label{tab: datasets}
\begin{tabular}{c|c|c|c|c}
\hline\hline
Dataset  & Type   & Samples & Classes & Edges \\ 
\hline\hline
USPS     & Image  & 9298    & 10      & 27894       \\
Reuters  & Text   & 10000   & 4       & 30000      \\
HHAR     & Record & 10299   & 6       & 30897       \\
\hline
ACM      & Graph  & 3025    & 3       & 13128      \\
CiteSeer & Graph  & 3327    & 6       & 4552      \\ 
DBLP     & Graph  & 4057    & 4       & 3528       \\ 
Amazon Photo     & Graph  & 7650    & 8       & 119081\\ 
PubMed     & Graph  & 19717    & 3       & 44324\\ 
AIDS     & Graph  & 31385    & 38       & 64780 \\ 
\hline\hline
\end{tabular}
\end{table}

% \begin{table}[]
% \begin{tabular}{cccccc}
% Dataset      & Type   & Samples & Edges  & Classes & Dimension \\
% USPS         & Image  & 9298    & 27894  & 10      & 256       \\
% Reuters      & Text   & 10000   & 30000  & 4       & 2000      \\
% HHAR         & Record & 10299   & 30897  & 6       & 561       \\
% ACM          & Graph  & 3025    & 26256  & 3       & 1870      \\
% CiteSeer     & Graph  & 3327    & 9104   & 6       & 3703      \\
% DBLP         & Graph  & 4057    & 7056   & 4       & 334       \\
% Amazon Photo & Graph  & 7650    & 238163 & 8       & 745       \\
% PubMed       & Graph  & 19717   & 88651  & 3       & 500       \\
% AIDS         & Graph  & 31385   & 64780  & 38      & 4        
% \end{tabular}
% \end{table}

\begin{table*}
\centering
\caption{Clustering results (mean$\pm$std) with twelve compared methods on nine benchmark datasets. The best and second-best results are \textbf{bolded} and \underline{underlined}, respectively. `OOM' denotes the Out Of Memory case.}
\label{tab: all_results}
\resizebox{1.01\textwidth}{!}{
\begin{tabular}{c|c|cccccccccccc|c}
\hline\hline
\multirow{2}{*}{Datasets}                        & \multirow{2}{*}{Metrics} & K-means \cite{macqueen1967some}     & DAE \cite{hinton2006reducing}     & DEC \cite{xie2016unsupervised}    & IDEC \cite{guo2017improved}       & GAE \cite{kipf2016variational}    & VGAE \cite{kipf2016variational}   & DAEGC \cite{wang2019attributed}               & ARGA \cite{pan2019learning}       & SDCN \cite{bo2020structural}      & AGCN \cite{peng2021attention}                 & DFCN \cite{tu2021deep}     &  \textcolor{black}{AGCC} \cite{he2022parallelly}            & Our                                        \\
                                                 &                          &                                                      & [Science06]                       & [ICML16]                          & [AAAI17]                          & [NIPS16]                          & [NIPS16]                          & [AAAI19]                                      & [IJCAI18]                         & [WWW20]                           & [MM21]                                        & [AAAI21]                   & [TNNLS22] &                                            \\
\hline\hline
\multirow{4}{*}{Reuters}                         & ARI                      & \textcolor{black}{46.09$\pm$0.02}   & 49.55$\pm$0.37                    & 48.44$\pm$0.14                    & 51.26$\pm$0.21                    & 19.61$\pm$0.22                    & 26.18$\pm$0.36                    & 31.12$\pm$0.18                                & 24.50$\pm$0.40                    & 55.36$\pm$0.37                    & 60.55$\pm$1.78                    & 59.80$\pm$0.40             & \underline{62.98$\pm$2.24} & \textbf{63.48$\pm$1.10}                    \\
                                                 & F1                       & \textcolor{black}{58.33$\pm$0.03}   & 60.96$\pm$0.22                    & 64.25$\pm$0.22                    & 63.21$\pm$0.12                    & 43.53$\pm$0.42                    & 57.14$\pm$0.17                    & 61.82$\pm$0.13                                & 51.10$\pm$0.20                    & 65.48$\pm$0.08                    & {66.16$\pm$0.64}                              & \textbf{69.60$\pm$0.10}    & 67.21$\pm$1.61 & \underline{68.81$\pm$1.26}                 \\
                                                 & ACC                      & \textcolor{black}{59.98$\pm$0.02}   & 74.90$\pm$0.21                    & 73.58$\pm$0.13                    & 75.43$\pm$0.14                    & 54.40$\pm$0.27                    & 60.85$\pm$0.23                    & 65.50$\pm$0.13                                & 56.20$\pm$0.20                    & 77.15$\pm$0.21                    & 79.30$\pm$1.07                    & 77.70$\pm$0.20             & \underline{81.65$\pm$1.52} & \textbf{81.68$\pm$0.69}                    \\
                                                 & NMI                      & \textcolor{black}{{58.86$\pm$0.01}} & 49.69$\pm$0.29                    & 47.50$\pm$0.34                    & 50.28$\pm$0.17                    & 25.92$\pm$0.41                    & 25.51$\pm$0.22                    & 30.55$\pm$0.29                                & 28.70$\pm$0.30                    & 50.82$\pm$0.21                    & {57.83$\pm$1.01}                              & \textbf{59.90$\pm$0.40}    & \underline{59.56$\pm$0.94} & 58.94$\pm$1.16                 \\
\hline\multirow{4}{*}{HHAR}                            & ARI                      & \textcolor{black}{27.95$\pm$0.38}   & 60.36$\pm$0.88                    & 61.25$\pm$0.51                    & 62.83$\pm$0.45                    & 42.63$\pm$1.63                    & 51.47$\pm$0.73                    & 60.38$\pm$2.15                                & 44.70$\pm$1.00                    & 72.84$\pm$0.09                    & \underline{77.07$\pm$0.66}                    & 76.40$\pm$0.10             & 75.58$\pm$1.85 & \textbf{77.38$\pm$0.97}                    \\
                                                 & F1                       & \textcolor{black}{41.28$\pm$2.43}   & 66.36$\pm$0.34                    & 67.29$\pm$0.29                    & 68.63$\pm$0.33                    & 62.64$\pm$0.97                    & 71.55$\pm$0.29                    & 76.89$\pm$2.18                                & 61.10$\pm$0.90                    & 82.58$\pm$0.08                    & \textbf{88.00$\pm$0.53}                       & 87.30$\pm$0.10             & 85.79$\pm$2.48 & \underline{87.90$\pm$1.11}                 \\
                                                 & ACC                      & \textcolor{black}{54.04$\pm$0.01}   & 68.69$\pm$0.31                    & 69.39$\pm$0.25                    & 71.05$\pm$0.36                    & 62.33$\pm$1.01                    & 71.30$\pm$0.36                    & 76.51$\pm$2.19                                & 63.30$\pm$0.80                    & 84.26$\pm$0.17                    & \textbf{88.11$\pm$0.43}                       & 87.10$\pm$0.10             & 86.54$\pm$1.79 & \underline{87.83$\pm$1.01}                 \\
                                                 & NMI                      & \textcolor{black}{41.54$\pm$0.51}   & 71.42$\pm$0.97                    & 72.91$\pm$0.39                    & 74.19$\pm$0.39                    & 55.06$\pm$1.39                    & 62.95$\pm$0.36                    & 69.10$\pm$2.28                                & 57.10$\pm$1.40                    & 79.90$\pm$0.09                    & \underline{82.44$\pm$0.62}                    & 82.20$\pm$0.10             & 82.21$\pm$1.78 & \textbf{85.34$\pm$2.11}                    \\
\hline\multirow{4}{*}{USPS}                            & ARI                      & \textcolor{black}{54.55$\pm$0.06}   & 58.83$\pm$0.05                    & 63.70$\pm$0.27                    & 67.86$\pm$0.12                    & 50.30$\pm$0.55                    & 40.96$\pm$0.59                    & 63.33$\pm$0.34                                & 51.10$\pm$0.60                    & 71.84$\pm$0.24                    & {73.61$\pm$0.43}                              & \underline{75.30$\pm$0.20} & 68.50$\pm$3.83 & \textbf{75.54$\pm$1.28}                    \\
                                                 & F1                       & \textcolor{black}{64.78$\pm$0.03}   & 69.74$\pm$0.03                    & 71.82$\pm$0.21                    & 74.63$\pm$0.10                    & 61.84$\pm$0.43                    & 53.63$\pm$1.05                    & 72.45$\pm$0.49                                & 66.10$\pm$1.20                    & 76.98$\pm$0.18                    & {77.61$\pm$0.38}                              & \underline{78.30$\pm$0.20} & 74.86$\pm$2.56 & \textbf{79.33$\pm$0.74}                    \\
                                                 & ACC                      & \textcolor{black}{66.82$\pm$0.04}   & 71.04$\pm$0.03                    & 73.31$\pm$0.17                    & 76.22$\pm$0.12                    & 63.10$\pm$0.33                    & 56.19$\pm$0.72                    & 73.55$\pm$0.40                                & 66.80$\pm$0.70                    & 78.08$\pm$0.19                    & \underline{80.98$\pm$0.28}                    & 79.50$\pm$0.20             & 77.14$\pm$1.21 & \textbf{81.13$\pm$1.89}                    \\
                                                 & NMI                      & \textcolor{black}{62.63$\pm$0.05}   & 67.53$\pm$0.03                    & 70.58$\pm$0.25                    & 75.56$\pm$0.06                    & 60.69$\pm$0.58                    & 51.08$\pm$0.37                    & 71.12$\pm$0.24                                & 61.60$\pm$0.30                    & 79.51$\pm$0.27                    & {79.64$\pm$0.32}                              & \textbf{82.80$\pm$0.30}    & 75.93$\pm$3.83 & \underline{82.14$\pm$0.15}                 \\
\hline\multirow{4}{*}{ACM}                             & ARI                      & \textcolor{black}{30.60$\pm$0.69}   & 54.64$\pm$0.16                    & 60.64$\pm$1.87                    & 62.16$\pm$1.50                    & 59.46$\pm$3.10                    & 57.72$\pm$0.67                    & 59.35$\pm$3.89                                & 62.90$\pm$2.10                    & 73.91$\pm$0.40                    & {74.20$\pm$0.38}                              & \underline{74.90$\pm$0.40} & 73.73$\pm$0.90 & \textbf{76.72$\pm$0.98}                    \\
                                                 & F1                       & \textcolor{black}{67.57$\pm$0.74}   & 82.01$\pm$0.08                    & 84.51$\pm$0.74                    & 85.11$\pm$0.48                    & 84.65$\pm$1.33                    & 84.17$\pm$0.23                    & 87.07$\pm$2.79                                & 86.10$\pm$1.20                    & 90.42$\pm$0.19                    & {90.58$\pm$0.17}                              & \underline{90.80$\pm$0.20} & 90.39$\pm$0.39 & \textbf{91.53$\pm$0.42}                    \\
                                                 & ACC                      & \textcolor{black}{67.31$\pm$0.71}   & 81.83$\pm$0.08                    & 84.33$\pm$0.76                    & 85.12$\pm$0.52                    & 84.52$\pm$1.44                    & 84.13$\pm$0.22                    & 86.94$\pm$2.83                                & 86.10$\pm$1.20                    & 90.45$\pm$0.18                    & {90.59$\pm$0.15}                              & \underline{90.90$\pm$0.20} & 90.38$\pm$0.38 & \textbf{91.55$\pm$0.40}                    \\
                                                 & NMI                      & \textcolor{black}{32.44$\pm$0.46}   & 49.30$\pm$0.16                    & 54.54$\pm$1.51                    & 56.61$\pm$1.16                    & 55.38$\pm$1.92                    & 53.20$\pm$0.52                    & 56.18$\pm$4.15                                & 55.70$\pm$1.40                    & 68.31$\pm$0.25                    & {68.38$\pm$0.45}                              & \underline{69.40$\pm$0.40} & 68.34$\pm$0.89 & \textbf{71.50$\pm$0.80}                    \\
\hline\multirow{4}{*}{CiteSeer}                        & ARI                      & \textcolor{black}{06.97$\pm$0.39}   & 29.31$\pm$0.14                    & 28.12$\pm$0.36                    & 25.70$\pm$2.65                    & 33.55$\pm$1.18                    & 33.13$\pm$0.53                    & 37.78$\pm$1.24                                & 33.40$\pm$1.50                    & 40.17$\pm$0.43                    & {43.79$\pm$0.31}                              & \underline{45.50$\pm$0.30} & 41.82$\pm$2.03 & \textbf{47.98$\pm$0.91}                    \\
                                                 & F1                       & \textcolor{black}{31.92$\pm$0.27}   & 53.80$\pm$0.11                    & 52.62$\pm$0.17                    & 61.62$\pm$1.39                    & 57.36$\pm$0.82                    & 57.70$\pm$0.49                    & 62.20$\pm$1.32                                & 54.80$\pm$0.80                    & \underline{63.62$\pm$0.24}        & 62.37$\pm$0.21                                & \textbf{64.30$\pm$0.20}    & 60.47$\pm$1.57 & {62.37$\pm$0.52}                           \\
                                                 & ACC                      & \textcolor{black}{38.65$\pm$0.65}   & 57.08$\pm$0.13                    & 55.89$\pm$0.20                    & 60.49$\pm$1.42                    & 61.35$\pm$0.80                    & 60.97$\pm$0.36                    & 64.54$\pm$1.39                                & 56.90$\pm$0.70                    & 65.96$\pm$0.31                    & {68.79$\pm$0.23}                              & \underline{69.50$\pm$0.20} & 68.08$\pm$1.44 & \textbf{72.01$\pm$0.53}                    \\
                                                 & NMI                      & \textcolor{black}{11.45$\pm$0.38}   & 27.64$\pm$0.08                    & 28.34$\pm$0.30                    & 27.17$\pm$2.40                    & 34.63$\pm$0.65                    & 32.69$\pm$0.27                    & 36.41$\pm$0.86                                & 34.50$\pm$0.80                    & 38.71$\pm$0.32                    & {41.54$\pm$0.30}                              & \underline{43.90$\pm$0.20} & 40.86$\pm$1.45 & \textbf{45.34$\pm$0.70}                    \\
\hline\multirow{4}{*}{DBLP}                            & ARI                      & \textcolor{black}{13.43$\pm$3.02}   & 12.21$\pm$0.43                    & 23.92$\pm$0.39                    & 25.37$\pm$0.60                    & 22.02$\pm$1.40                    & 17.92$\pm$0.07                    & 21.03$\pm$0.52                                & 22.70$\pm$0.30                    & 39.15$\pm$2.01                    & {42.49$\pm$0.31}                              & \underline{47.00$\pm$1.50} & 44.40$\pm$3.79 & \textbf{57.29$\pm$1.20}                    \\
                                                 & F1                       & \textcolor{black}{36.08$\pm$3.53}   & 52.53$\pm$0.36                    & 59.38$\pm$0.51                    & 61.33$\pm$0.56                    & 61.41$\pm$2.23                    & 58.69$\pm$0.07                    & 61.75$\pm$0.67                                & 61.80$\pm$0.90                    & 67.71$\pm$1.51                    & {72.80$\pm$0.56}                              & \underline{75.70$\pm$0.80} & 71.84$\pm$2.02 & \textbf{80.79$\pm$0.61}                    \\
                                                 & ACC                      & \textcolor{black}{39.32$\pm$3.17}   & 51.43$\pm$0.35                    & 58.16$\pm$0.56                    & 60.31$\pm$0.62                    & 61.21$\pm$1.22                    & 58.59$\pm$0.06                    & 62.05$\pm$0.48                                & 61.60$\pm$1.00                    & 68.05$\pm$1.81                    & {73.26$\pm$0.37}                              & \underline{76.00$\pm$0.80} & 73.45$\pm$2.16 & \textbf{81.26$\pm$0.62}                    \\
                                                 & NMI                      & \textcolor{black}{16.94$\pm$3.22}   & 25.40$\pm$0.16                    & 29.51$\pm$0.28                    & 31.17$\pm$0.50                    & 30.80$\pm$0.91                    & 26.92$\pm$0.06                    & 32.49$\pm$0.45                                & 26.80$\pm$1.00                    & 39.50$\pm$1.34                    & {39.68$\pm$0.42}                              & \underline{43.70$\pm$1.00} & 40.36$\pm$2.81 & \textbf{51.99$\pm$0.76}                    \\
\hline\textcolor{black}{\multirow{4}{*}{Amazon Photo}} & ARI                      & \textcolor{black}{05.50$\pm$0.44}   & \textcolor{black}{20.80$\pm$0.47} & \textcolor{black}{18.59$\pm$0.04} & \textcolor{black}{19.24$\pm$0.07} & \textcolor{black}{48.82$\pm$4.57} & \textcolor{black}{56.24$\pm$4.66} & \textcolor{black}{\underline{59.39$\pm$0.02}} & \textcolor{black}{44.18$\pm$4.41} & \textcolor{black}{31.21$\pm$1.23} & {41.15$\pm$2.78}                              & 58.98$\pm$0.84             & 29.96$\pm$3.46
             & \textcolor{black}{\textbf{60.51$\pm$1.58}} \\
                                                 & F1                       & \textcolor{black}{23.96$\pm$0.51}   & \textcolor{black}{47.87$\pm$0.20} & \textcolor{black}{46.71$\pm$0.12} & \textcolor{black}{47.20$\pm$0.11} & \textcolor{black}{68.08$\pm$1.76} & {70.38$\pm$2.98}                  & \textcolor{black}{69.97$\pm$0.02}             & \textcolor{black}{64.30$\pm$1.95} & \textcolor{black}{50.66$\pm$1.49} & \textcolor{black}{43.68$\pm$5.08}             & \underline{71.58$\pm$0.31} & 39.67$\pm$5.22
             & \textcolor{black}{\textbf{71.68$\pm$2.35}} \\
                                                 & ACC                      & \textcolor{black}{27.22$\pm$0.76}   & \textcolor{black}{48.25$\pm$0.08} & \textcolor{black}{47.22$\pm$0.08} & \textcolor{black}{47.62$\pm$0.08} & \textcolor{black}{71.57$\pm$2.48} & \textcolor{black}{74.26$\pm$3.63} & {76.44$\pm$0.01}                              & \textcolor{black}{69.28$\pm$2.30} & \textcolor{black}{53.44$\pm$0.81} & \textcolor{black}{58.53$\pm$1.74}             & \underline{76.88$\pm$0.80} &  51.47$\pm$3.04
            & \textcolor{black}{\textbf{78.75$\pm$1.02}} \\
                                                 & NMI                      & \textcolor{black}{13.23$\pm$1.33}   & \textcolor{black}{38.76$\pm$0.30} & \textcolor{black}{37.35$\pm$0.05} & \textcolor{black}{37.83$\pm$0.08} & \textcolor{black}{62.13$\pm$2.79} & {66.01$\pm$3.40}                  & \textcolor{black}{65.57$\pm$0.03}             & \textcolor{black}{58.36$\pm$2.76} & \textcolor{black}{44.85$\pm$0.83} & \textcolor{black}{51.76$\pm$3.23}             & \textbf{69.21$\pm$1.00}    &  39.19$\pm$4.07
            & \underline{66.27$\pm$1.13}                 \\
\hline\textcolor{black}{\multirow{4}{*}{PubMed}}       & ARI                      & \textcolor{black}{28.10$\pm$0.01}   & \textcolor{black}{23.86$\pm$0.67} & \textcolor{black}{19.55$\pm$0.13} & \textcolor{black}{20.58$\pm$0.39} & \textcolor{black}{20.62$\pm$1.39} & \textcolor{black}{30.15$\pm$1.23} & \textcolor{black}{29.84$\pm$0.04}             & \textcolor{black}{24.35$\pm$0.17} & \textcolor{black}{22.30$\pm$2.07} & \textcolor{black}{\underline{31.39$\pm$0.67}} & 30.64$\pm$0.11             & OOM             & \textcolor{black}{\textbf{35.29$\pm$1.02}} \\
                                                 & F1                       & \textcolor{black}{58.88$\pm$0.01}   & \textcolor{black}{64.01$\pm$0.29} & \textcolor{black}{61.49$\pm$0.10} & \textcolor{black}{62.41$\pm$0.32} & \textcolor{black}{61.37$\pm$0.85} & \textcolor{black}{67.68$\pm$0.89} & \textcolor{black}{68.23$\pm$0.02}             & \textcolor{black}{65.69$\pm$0.13} & \textcolor{black}{65.01$\pm$1.21} & \textcolor{black}{\underline{69.73$\pm$0.45}} & 68.10$\pm$0.07             & OOM             & \textcolor{black}{\textbf{72.78$\pm$0.72}} \\
                                                 & ACC                      & \textcolor{black}{59.83$\pm$0.01}   & \textcolor{black}{63.07$\pm$0.31} & \textcolor{black}{60.14$\pm$0.09} & \textcolor{black}{60.70$\pm$0.34} & \textcolor{black}{62.09$\pm$0.81} & \textcolor{black}{68.48$\pm$0.77} & \textcolor{black}{68.73$\pm$0.03}             & \textcolor{black}{65.26$\pm$0.12} & \textcolor{black}{64.20$\pm$1.30} & \textcolor{black}{\underline{69.67$\pm$0.42}} & 68.89$\pm$0.07             & OOM             & \textcolor{black}{\textbf{73.16$\pm$0.69}} \\
                                                 & NMI                      & {31.05$\pm$0.02}                    & \textcolor{black}{26.32$\pm$0.57} & \textcolor{black}{22.44$\pm$0.14} & \textcolor{black}{23.67$\pm$0.29} & \textcolor{black}{23.84$\pm$3.54} & \textcolor{black}{30.61$\pm$1.71} & \textcolor{black}{28.26$\pm$0.03}             & \textcolor{black}{24.80$\pm$0.17} & \textcolor{black}{22.87$\pm$2.04} & \textcolor{black}{30.96$\pm$0.99}             & \underline{31.43$\pm$0.13} &  OOM            & \textcolor{black}{\textbf{33.29$\pm$1.14}} \\
\hline\multirow{4}{*}{\textcolor{black}{AIDS}} & ARI     & 05.37$\pm$0.19                  & 05.71$\pm$0.66                & 10.71$\pm$3.49                 & 13.39$\pm$5.35              & 03.50$\pm$0.79                  & 00.44$\pm$0.43                   & OOM                             & 01.79$\pm$0.97               & $-$0.06$\pm$0.00             & \underline{14.03$\pm$5.76}          & 00.48$\pm$0.00          & OOM                          & \textbf{21.40$\pm$7.12} \\
                      & F1      & 11.86$\pm$1.02                  & 11.91$\pm$1.54                & \underline{13.81$\pm$1.60}           & 12.10$\pm$1.55              & 08.40$\pm$1.23                  & 09.13$\pm$2.74                   & OOM                             & 06.01$\pm$1.41               & 02.02$\pm$0.00                & 06.14$\pm$1.80                & 05.51$\pm$0.01          & OOM                          & \textbf{21.77$\pm$1.10} \\
                      & ACC     & 17.11$\pm$0.63                  & 21.78$\pm$5.02                & 35.12$\pm$3.69                 & 47.32$\pm$5.76              & 15.72$\pm$0.89                 & 23.04$\pm$6.08                  & OOM                             & 59.27$\pm$2.54              & \underline{62.25$\pm$0.00}         & 59.82$\pm$3.37                & 11.28$\pm$0.02         & OOM                          & \textbf{63.84$\pm$2.81} \\
                      & NMI     & 23.42$\pm$0.57                  & 24.29$\pm$2.49                & 24.30$\pm$2.19                 & \underline{25.37$\pm$4.90}        & 12.66$\pm$2.64                 & 01.62$\pm$0.29                   & OOM                             & 04.86$\pm$2.16               & 00.16$\pm$0.00                & 09.08$\pm$2.28                & 04.67$\pm$0.01          & OOM                          & \textbf{34.44$\pm$2.96}\\
\hline\hline
\end{tabular}
}
\end{table*}

\section{Experiments}\label{sec: eprm}
We conducted quantitative and qualitative experiments on nine commonly used benchmark datasets to evaluate the proposed model. In addition, we performed ablation studies to investigate the effectiveness of the proposed modules and the adopted strategies. Moreover, we performed a series of parameter analyses to verify the robustness of our method. 

\subsection{Datasets and Compared Methods}
We conducted experiments on one image dataset (\textbf{USPS} \cite{hull1994database}), one text dataset (\textbf{Reuters} \cite{lewis2004rcv1}), one record dataset (\textbf{HHAR} \cite{stisen2015smart}), and six graph datasets (\textbf{ACM}$\footnote{http://dl.acm.org}$, \textbf{CiteSeer}$\footnote{http://CiteSeerx.ist.psu.edu/}$, \textbf{DBLP}$\footnote{https://dblp.uni-trier.de}$, \textbf{Amazon Photo}, \textbf{PubMed}\cite{wan2021contrastive}, and \textbf{AIDS}\cite{riesen2008iam}), which are briefly summarized in Table \ref{tab: datasets}. 

We compared the proposed method with the classic clustering method K-means \cite{macqueen1967some}, three DAE-based embedding clustering methods \cite{hinton2006reducing,xie2016unsupervised,guo2017improved}, and seven GCN-based embedding clustering methods \cite{kipf2016variational,pan2019learning,wang2019attributed,velivckovic2018graph,bo2020structural,peng2021attention,tu2021deep}, the details of which are listed as follows.

\begin{table*}[]
\centering
\caption{Results of ablation studies. \xmark \ and \cmark \ indicate the component is used and unused, respectively. The best results are noted in \textbf{bold}.}
\label{tab: AS}
\resizebox{0.78\textwidth}{!}{%{0.57\textwidth}{!}{
\begin{tabular}{c|ccccc|cccc}
\hline\hline
Datasets                  & SSS                            & HSS                            & DWF                            & SWF                            & HWF                            & ARI                                       & F1                                        & ACC                                       & NMI                                       \\
\hline\multirow{7}{*}{USPS}     &  \xmark                              & \xmark                               & \xmark                               & \xmark                               & \xmark                               & 71.67$\pm$0.44                            & 76.88$\pm$0.30                            & 78.08$\pm$0.30                            & 79.19$\pm$0.44                            \\
                          &  \xmark                              & \xmark                               & \xmark                               & \xmark                               & \textcolor{black}{\cmark} & \textcolor{black}{71.71$\pm$0.87}          & \textcolor{black}{76.46$\pm$0.54}          & \textcolor{black}{78.98$\pm$0.97}          & \textcolor{black}{78.87$\pm$0.36}          \\
                         &  \xmark                              & \xmark                               & \xmark                                & \cmark                   & \cmark                   & 70.96$\pm$0.24                            & 76.44$\pm$0.17                            & 77.70$\pm$0.14                            & 78.61$\pm$0.22                            \\
                          &  \xmark                              & \xmark                                & \textcolor{black}{\cmark} & \textcolor{black}{\cmark} & \textcolor{black}{\cmark} & \textcolor{black}{71.73$\pm$0.71}          & \textcolor{black}{76.31$\pm$0.34}          & \textcolor{black}{79.63$\pm$0.43}          & \textcolor{black}{78.41$\pm$0.29}          \\%\cline{2-10}
                          &  \xmark                                & \textcolor{black}{\cmark} & \textcolor{black}{\cmark} & \textcolor{black}{\cmark} & \textcolor{black}{\cmark} & \textcolor{black}{71.90$\pm$0.99}          & \textcolor{black}{76.61$\pm$0.56}          & \textcolor{black}{79.74$\pm$0.79}          & \textcolor{black}{78.64$\pm$0.46}          \\ 
                          & \cmark                   &  \xmark                                & \cmark                   & \cmark                   & \cmark                   & 74.39$\pm$0.13                            & 78.51$\pm$0.09                            & 79.11$\pm$0.11                            & 82.06$\pm$0.16                            \\
                          & \cmark                   & \cmark                   & \cmark                   & \cmark                   & \cmark                   & \textbf{75.54$\pm$1.28}                   & \textbf{79.33$\pm$0.74}                   & \textbf{81.13$\pm$1.89}                   & \textbf{82.14$\pm$0.15}                   \\
\hline\multirow{7}{*}{Reuters}  &  \xmark                              & \xmark                               & \xmark                               & \xmark                               & \xmark                                & 56.37$\pm$4.76                            & 65.03$\pm$1.87                            & 78.19$\pm$2.02                            & 53.74$\pm$3.63                            \\
                          &  \xmark&  \xmark&  \xmark&  \xmark                                & \textcolor{black}{\cmark} & \textcolor{black}{61.38$\pm$0.78}          & \textcolor{black}{67.22$\pm$1.15}          & \textcolor{black}{80.19$\pm$0.53}          & \textcolor{black}{57.94$\pm$0.49}          \\
                          &  \xmark&  \xmark&  \xmark                                & \cmark                   & \cmark                   & 61.55$\pm$0.64                            & 66.54$\pm$0.21                            & 80.60$\pm$0.47                            & 58.15$\pm$0.49                            \\
                          &  \xmark&  \xmark                                & \textcolor{black}{\cmark} & \textcolor{black}{\cmark} & \textcolor{black}{\cmark} & \textcolor{black}{62.70$\pm$1.00}          & \textcolor{black}{66.90$\pm$0.30}          & \textcolor{black}{80.95$\pm$0.46}          & \textcolor{black}{59.42$\pm$0.69}          \\
                          &  \xmark                                & \textcolor{black}{\cmark} & \textcolor{black}{\cmark} & \textcolor{black}{\cmark} & \textcolor{black}{\cmark} & \textcolor{black}{63.32$\pm$0.57}          & \textcolor{black}{67.21$\pm$0.18}          & \textcolor{black}{81.28$\pm$0.32}          & \textbf{\textcolor{black}{60.79$\pm$0.69}} \\
                          & \cmark                   &  \xmark                                & \cmark                   & \cmark                   & \cmark                   & 62.75$\pm$2.00                            & 68.74$\pm$1.23                            & 81.02$\pm$0.81                            & 57.93$\pm$1.50                            \\
                          & \cmark                   & \cmark                   & \cmark                   & \cmark                   & \cmark                   & \textbf{63.48$\pm$1.10}                   & \textbf{68.81$\pm$1.26}                   & \textbf{81.68$\pm$0.69}                   & 58.94$\pm$1.16                            \\
\hline\multirow{7}{*}{HHAR}     &  \xmark                              & \xmark                               & \xmark                               & \xmark                               & \xmark                               & 73.17$\pm$1.95                            & 82.70$\pm$3.97                            & 84.18$\pm$2.80                            & 80.03$\pm$1.16                            \\
                          &  \xmark&  \xmark&  \xmark&  \xmark                               & \textcolor{black}{\cmark} & \textcolor{black}{72.45$\pm$1.02}          & \textcolor{black}{83.25$\pm$0.81}          & \textcolor{black}{84.60$\pm$0.66}          & \textcolor{black}{79.08$\pm$0.85}          \\
                          &  \xmark&  \xmark&  \xmark                                & \cmark                   & \cmark                   & 73.24$\pm$0.73                            & 83.34$\pm$1.69                            & 84.77$\pm$1.21                            & 80.10$\pm$0.50                            \\
                          &  \xmark&  \xmark                              & \textcolor{black}{\cmark} & \textcolor{black}{\cmark} & \textcolor{black}{\cmark} & \textcolor{black}{72.84$\pm$1.23}          & \textcolor{black}{83.72$\pm$1.10}          & \textcolor{black}{84.95$\pm$0.86}          & \textcolor{black}{79.22$\pm$0.94}          \\
                         &  \xmark                               & \textcolor{black}{\cmark} & \textcolor{black}{\cmark} & \textcolor{black}{\cmark} & \textcolor{black}{\cmark} & \textcolor{black}{73.24$\pm$0.52}          & \textcolor{black}{83.74$\pm$0.68}          & \textcolor{black}{85.01$\pm$0.46}          & \textcolor{black}{79.99$\pm$0.47}          \\
                          & \cmark                   &  \xmark                                & \cmark                   & \cmark                   & \cmark                   & 75.91$\pm$0.40                            & 86.65$\pm$0.70                            & 86.23$\pm$0.81                            & 82.61$\pm$0.12                            \\
                          & \cmark                   & \cmark                   & \cmark                   & \cmark                   & \cmark                   & \textbf{77.38$\pm$0.97}                   & \textbf{87.90$\pm$1.11}                   & \textbf{87.83$\pm$1.01}                   & \textbf{85.34$\pm$2.11}                   \\
\hline\multirow{7}{*}{ACM}      &  \xmark                              & \xmark                               & \xmark                               & \xmark                               & \xmark                               & 73.91$\pm$0.40                            & 90.42$\pm$0.19                            & 90.45$\pm$0.18                            & 68.31$\pm$0.25                            \\
                          &  \xmark&  \xmark&  \xmark&  \xmark                               & \textcolor{black}{\cmark} & \textcolor{black}{73.95$\pm$0.60}          & \textcolor{black}{90.48$\pm$0.26}          & \textcolor{black}{90.47$\pm$0.24}          & \textcolor{black}{68.42$\pm$0.61}          \\
                         &  \xmark&  \xmark&  \xmark                               & \cmark                   & \cmark                   & 74.20$\pm$0.38                            & 90.58$\pm$0.17                            & 90.59$\pm$0.15                            & 68.38$\pm$0.45                            \\
                         &  \xmark&  \xmark                               & \textcolor{black}{\cmark} & \textcolor{black}{\cmark} & \textcolor{black}{\cmark} & \textcolor{black}{74.58$\pm$0.78}          & \textcolor{black}{90.72$\pm$0.35}          & \textcolor{black}{90.73$\pm$0.33}          & \textcolor{black}{68.94$\pm$0.63}          \\
                          &  \xmark                                & \textcolor{black}{\cmark} & \textcolor{black}{\cmark} & \textcolor{black}{\cmark} & \textcolor{black}{\cmark} & \textcolor{black}{74.83$\pm$0.73}          & \textcolor{black}{90.85$\pm$0.33}          & \textcolor{black}{90.85$\pm$0.31}          & \textcolor{black}{69.02$\pm$0.66}          \\
                          & \cmark                   &  \xmark                                & \cmark                   & \cmark                   & \cmark                   & 75.78$\pm$0.64                            & 91.18$\pm$0.25                            & 91.18$\pm$0.26                            & 70.59$\pm$0.68                            \\
                          & \cmark                   & \cmark                   & \cmark                   & \cmark                   & \cmark                   & \textbf{76.72$\pm$0.98}                   & \textbf{91.53$\pm$0.42}                   & \textbf{91.55$\pm$0.40}                   & \textbf{71.50$\pm$0.80}                   \\
\hline\multirow{7}{*}{CiteSeer} &  \xmark                              & \xmark                               & \xmark                               & \xmark                               & \xmark                                & 40.17$\pm$0.43                            & \textbf{63.62$\pm$0.24}                   & 65.96$\pm$0.31                            & 38.71$\pm$0.32                            \\
                         &  \xmark&  \xmark&  \xmark&  \xmark                               & \textcolor{black}{\cmark} & \textcolor{black}{40.93$\pm$1.78}          & \textcolor{black}{60.91$\pm$0.81}          & \textcolor{black}{66.38$\pm$1.72}          & \textcolor{black}{39.07$\pm$1.52}          \\
                          &  \xmark&  \xmark&  \xmark                                & \cmark                   & \cmark                   & 43.79$\pm$0.31                            & 62.37$\pm$0.21                            & 68.79$\pm$0.23                            & 41.54$\pm$0.30                            \\
                          &  \xmark&  \xmark                               & \textcolor{black}{\cmark} & \textcolor{black}{\cmark} & \textcolor{black}{\cmark} & \textcolor{black}{43.50$\pm$0.47}          & \textcolor{black}{61.25$\pm$0.31}          & \textcolor{black}{68.54$\pm$0.30}          & \textcolor{black}{41.35$\pm$0.58}          \\
                          &  \xmark                               & \textcolor{black}{\cmark} & \textcolor{black}{\cmark} & \textcolor{black}{\cmark} & \textcolor{black}{\cmark} & \textcolor{black}{43.72$\pm$0.60}          & \textcolor{black}{61.52$\pm$0.65}          & \textcolor{black}{68.46$\pm$0.40}          & \textcolor{black}{41.25$\pm$0.41}          \\
                          & \cmark                   &  \xmark                                & \cmark                   & \cmark                   & \cmark                   & 47.76$\pm$1.28                            & 62.24$\pm$0.80                            & 71.86$\pm$0.79                            & 45.10$\pm$1.05                            \\
                          & \cmark                   & \cmark                   & \cmark                   & \cmark                   & \cmark                   & \textbf{47.98$\pm$0.91}                   & 62.37$\pm$0.52                            & \textbf{72.01$\pm$0.53}                   & \textbf{45.34$\pm$0.70}                   \\
\hline\multirow{7}{*}{DBLP}     &  \xmark                              & \xmark                               & \xmark                               & \xmark                               & \xmark                                & 39.15$\pm$2.01                            & 67.71$\pm$1.51                            & 68.05$\pm$1.81                            & 39.50$\pm$1.34                            \\
                          &  \xmark&  \xmark&  \xmark&  \xmark                                & \textcolor{black}{\cmark} & \textcolor{black}{37.78$\pm$1.85}          & \textcolor{black}{68.69$\pm$1.65}          & \textcolor{black}{69.65$\pm$1.43}          & \textcolor{black}{35.37$\pm$1.58}          \\
                          &  \xmark&  \xmark&  \xmark                                & \cmark                   & \cmark                   & 42.49$\pm$0.31                            & 72.80$\pm$0.56                            & 73.26$\pm$0.37                            & 39.68$\pm$0.42                            \\
                          &  \xmark&  \xmark                              & \textcolor{black}{\cmark} & \textcolor{black}{\cmark} & \textcolor{black}{\cmark} & \textcolor{black}{41.72$\pm$0.47}          & \textcolor{black}{72.68$\pm$0.20}          & \textcolor{black}{72.92$\pm$0.21}          & \textcolor{black}{39.26$\pm$0.33}          \\
                         &  \xmark                                & \textcolor{black}{\cmark} & \textcolor{black}{\cmark} & \textcolor{black}{\cmark} & \textcolor{black}{\cmark} & \textcolor{black}{42.52$\pm$0.96}          & \textcolor{black}{72.81$\pm$0.59}          & \textcolor{black}{73.43$\pm$0.50}          & \textcolor{black}{39.99$\pm$0.70}          \\
                          & \cmark                   &  \xmark                                & \cmark                   & \cmark                   & \cmark                   & 55.45$\pm$0.60                            & 79.83$\pm$0.32                            & 80.29$\pm$0.33                            & 50.08$\pm$0.56                            \\
                          & \cmark                   & \cmark                   & \cmark                   & \cmark                   & \cmark                   & \textbf{57.29$\pm$1.20}                   & \textbf{80.79$\pm$0.61}                   & \textbf{81.26$\pm$0.62}                   & \textbf{51.99$\pm$0.76}                   \\
\hline\multirow{7}{*}{\textcolor{black}{Amazon Photo}}    &  \xmark                              & \xmark                               & \xmark                               & \xmark                               & \xmark                              &  \textcolor{black}{31.21$\pm$1.23}          & \textcolor{black}{50.66$\pm$1.49}          & \textcolor{black}{53.44$\pm$0.81}          & \textcolor{black}{44.85$\pm$0.83}         \\
                         &  \xmark&  \xmark&  \xmark&  \xmark                                & \textcolor{black}{\cmark} & \textcolor{black}{37.86$\pm$3.46}          & \textcolor{black}{35.86$\pm$4.00}          & \textcolor{black}{54.84$\pm$1.43}          & \textcolor{black}{46.51$\pm$4.93}        \\
                         &  \xmark&  \xmark&  \xmark                               & \textcolor{black}{\cmark} & \textcolor{black}{\cmark} & \textcolor{black}{41.15$\pm$2.78}          & \textcolor{black}{43.68$\pm$5.08}          & \textcolor{black}{58.53$\pm$1.74}          & \textcolor{black}{51.76$\pm$3.23}      \\
                          &  \xmark&  \xmark                                & \textcolor{black}{\cmark} & \textcolor{black}{\cmark} & \textcolor{black}{\cmark} &  \textcolor{black}{41.14$\pm$2.78}          & \textcolor{black}{43.68$\pm$5.08}          & \textcolor{black}{58.52$\pm$1.74}          & \textcolor{black}{51.77$\pm$3.22}        \\
                          &  \xmark                                & \textcolor{black}{\cmark} & \textcolor{black}{\cmark} & \textcolor{black}{\cmark} & \textcolor{black}{\cmark} & \textcolor{black}{43.50$\pm$2.29}          & \textcolor{black}{46.20$\pm$4.18}          & \textcolor{black}{60.59$\pm$1.94}          & \textcolor{black}{52.23$\pm$1.67}        \\
                          & \textcolor{black}{\cmark}   &  \xmark         & \textcolor{black}{\cmark} & \textcolor{black}{\cmark} & \textcolor{black}{\cmark} & \textcolor{black}{51.81$\pm$2.25}          & \textcolor{black}{66.37$\pm$2.64}          & \textcolor{black}{71.93$\pm$2.08}          & \textcolor{black}{59.09$\pm$1.60}      \\
                          & \textcolor{black}{\cmark} & \textcolor{black}{\cmark} & \textcolor{black}{\cmark} & \textcolor{black}{\cmark} & \textcolor{black}{\cmark} & \textbf{\textcolor{black}{60.51$\pm$1.58}} & \textbf{\textcolor{black}{71.68$\pm$2.35}} & \textbf{\textcolor{black}{78.75$\pm$1.02}} & \textbf{\textcolor{black}{66.27$\pm$1.13}} 
                          \\
\hline\multirow{7}{*}{\textcolor{black}{PubMed}}  &  \xmark                              & \xmark                               & \xmark                               & \xmark                               & \xmark                               & \textcolor{black}{22.30$\pm$2.07}          & \textcolor{black}{65.01$\pm$1.21}          & \textcolor{black}{64.20$\pm$1.30}          & \textcolor{black}{22.87$\pm$2.04}          \\
                          &  \xmark&  \xmark&  \xmark&  \xmark                                & \textcolor{black}{\cmark}& \textcolor{black}{27.65$\pm$1.16}          & \textcolor{black}{67.21$\pm$0.83}          & \textcolor{black}{67.31$\pm$0.78}          & \textcolor{black}{27.77$\pm$1.85}        \\
                          &  \xmark&  \xmark&  \xmark                                & \textcolor{black}{\cmark} & \textcolor{black}{\cmark} &\textcolor{black}{31.39$\pm$0.67} & \textcolor{black}{69.73$\pm$0.45} & \textcolor{black}{69.67$\pm$0.42} & \textcolor{black}{30.96$\pm$0.99}        \\
                          &  \xmark&  \xmark                               & \textcolor{black}{\cmark} & \textcolor{black}{\cmark} & \textcolor{black}{\cmark}&    \textcolor{black}{30.85$\pm$1.10} & \textcolor{black}{69.05$\pm$0.87} & \textcolor{black}{68.67$\pm$0.79} & \textcolor{black}{32.19$\pm$1.29}        \\
                          &  \xmark                                & \textcolor{black}{\cmark} & \textcolor{black}{\cmark} & \textcolor{black}{\cmark} & \textcolor{black}{\cmark}& \textcolor{black}{33.21$\pm$1.94}          & \textcolor{black}{70.75$\pm$1.28}          & \textcolor{black}{71.35$\pm$1.39}          & \textcolor{black}{31.47$\pm$1.75}     \\
                          & \textcolor{black}{\cmark}    &  \xmark         & \textcolor{black}{\cmark} & \textcolor{black}{\cmark} & \textcolor{black}{\cmark}&\textcolor{black}{32.79$\pm$1.57}          & \textcolor{black}{69.89$\pm$1.20}          & \textcolor{black}{70.56$\pm$1.37}          & \textcolor{black}{31.85$\pm$1.36}      \\
                          & \textcolor{black}{\cmark} & \textcolor{black}{\cmark} & \textcolor{black}{\cmark} & \textcolor{black}{\cmark} & \textcolor{black}{\cmark} & \textbf{\textcolor{black}{35.29$\pm$1.02}} & \textbf{\textcolor{black}{72.78$\pm$0.72}} & \textbf{\textcolor{black}{73.16$\pm$0.69}} & \textbf{\textcolor{black}{33.29$\pm$1.14}} \\
\hline\multirow{7}{*}{\textcolor{black}{AIDS}} &  \xmark                              & \xmark                               & \xmark                               & \xmark                               & \xmark              & 10.15$\pm$3.04 & 4.76$\pm$0.69  & 58.32$\pm$3.33 & 7.67$\pm$0.48  \\
                      &  \xmark                              & \xmark                               & \xmark                               & \xmark                               & \cmark & 10.31$\pm$3.52 & 3.92$\pm$0.67  & 62.25$\pm$0.01 & 6.68$\pm$1.13  \\
                      &  \xmark                              & \xmark                               & \xmark                               & \cmark                               & \cmark & 10.69$\pm$3.33 & 4.33$\pm$0.99  & 62.32$\pm$0.21 & 7.54$\pm$1.40  \\
                      &  \xmark                              & \xmark                               & \cmark                               & \cmark                               & \cmark & 11.85$\pm$3.35 & 21.19$\pm$1.39 & 62.29$\pm$0.11 & 32.31$\pm$0.40 \\
                      &  \xmark                              & \cmark                               & \cmark                               & \cmark                               & \cmark & 12.78$\pm$2.61 & 21.65$\pm$1.51 & 62.25$\pm$0.00 & 32.34$\pm$0.35 \\
                      &  \cmark                              & \xmark                               & \cmark                               & \cmark                               & \cmark & 15.34$\pm$5.94 & 21.37$\pm$1.60 & 62.61$\pm$1.15 & 32.46$\pm$1.01 \\
                      &  \cmark                              & \cmark                               & \cmark                               & \cmark                               & \cmark & \textbf{21.40$\pm$7.12} & \textbf{21.77$\pm$1.10} & \textbf{63.84$\pm$2.81} & \textbf{34.44$\pm$2.96} \\
\hline\hline
\end{tabular}
}
\end{table*}

\begin{itemize}
	\item	\textbf{DAE} \cite{hinton2006reducing} uses deep auto-encoder to learn latent feature representations and then performs K-means on that feature to obtain clustering results.
	\item	\textbf{DEC} \cite{xie2016unsupervised} jointly conducts embedding learning and cluster assignment with an iterative procedure.
	\item	\textbf{IDEC} \cite{guo2017improved} introduces a reconstruction loss into DEC to improve the clustering performance.
	\item	\textbf{GAE} \cite{kipf2016variational} and \textbf{VGAE} \cite{kipf2016variational} incorporate DAE and variational DAE into GCN frameworks, respectively. 
	\item	\textbf{DAEGC} \cite{wang2019attributed} achieves a neighbor-wise embedding learning with an attention-driven strategy and supervises the network training with a clustering loss.
	\item	\textbf{ARGA} \cite{pan2019learning} guides embedding learning with a designed adversarial regularization.
	\item	\textbf{SDCN} \cite{bo2020structural} fuses DEC and GCN to merge the topological structure information into deep embedding clustering.
	\item  \textcolor{black}{\textbf{AGCN} \cite{peng2021attention} focuses on enhancing the embedding learning.}
	\item  \textbf{DFCN} \cite{tu2021deep} merges the node attribute and topological structure information based on the DAE and GAE.
	\item \textcolor{black}{\textbf{AGCC} \cite{he2022parallelly} replaces the graph layer by layer to mine the latent connected relationship between data.}
\end{itemize}

\subsection{Implementation Details}

\subsubsection{Evaluation metrics} 
We used four metrics to evaluate the clustering performance, including Average Rand Index (ARI), macro F1-score (F1), Accuracy (ACC), and Normalized Mutual Information (NMI). For each metric, a larger value implies a better clustering result.

\subsubsection{Graph construction} 
As those non-graph datasets (i.e., USPS, Reuters, and HHAR) lack the topology graph, we used a typical graph construction approach to generate their graph data. Specifically, we first employed the cosine distance to compute the similarity matrix $\mathbf{S}$, i.e.,
\begin{equation}
\begin{aligned}
&\mathbf{S}=\frac{\mathbf{X}\mathbf{X}^\mathsf{T}}{\left\|\mathbf{X}\right\|_F\left\|\mathbf{X}^\mathsf{T}\right\|_F},
\label{eq: Cosine}
\end{aligned}
\end{equation}
where $\left\|\mathbf{X}\right\|_F=\sqrt{\sum_{i=1}^{n}\sum_{j=1}^{d}{\left| \emph{x}_{i,j} \right|}^2}$ and $\mathbf{X}^\mathsf{T}$ denote the Frobenius norm and the transpose operation of $\mathbf{X}$, respectively. Then, we keep the top-$\hat{\emph{k}}$ similar neighbors of each sample to construct an undirected $\hat{\emph{k}}$-nearest neighbor (KNN \cite{altman1992introduction}) graph. The constructed KNN graph can depict the topological structure of a dataset and hence is used as GCN input.

\subsubsection{Training Procedure} 
Similar to \cite{xie2016unsupervised,guo2017improved,bo2020structural,peng2021attention}, we first pre-trained the DAE module with $30$ epochs and the learning rate equal to $0.001$. Then, we trained the whole network with $200$ iterations. We set the dimension of the auto-encoder and the GCN layers to $500-500-2000-10$, the batch size to $256$, and the negative input slope of LReLU to $0.2$. In addition, we set the learning rates of USPS, HHAR, ACM, DBLP, and \textcolor{black}{PubMed} datasets with $0.001$, and Reuters, CiteSeer, and \textcolor{black}{Amazon Photo} datasets with $0.0001$. \textcolor{black}{We set $r$ to $0.8$ in this paper, where more detailed experiments and analyses of the threshold value are given in Section \uppercase\expandafter{\romannumeral4}. E. 3).} For the method ARGA, we used the parameter settings given by the original paper \cite{pan2019learning}. For other methods under comparison, we directly cited the results in \cite{peng2021attention}. We repeated the experiment 10 times to evaluate our method with the mean values and the corresponding standard deviations (i.e., mean$\pm$std). The training procedure is implemented by PyTorch on two GPUs (GeForce RTX 2080 Ti and NVIDIA GeForce RTX 3090).

\subsection{Clustering Results}
Table \ref{tab: all_results} provides the clustering results of the proposed method and twelve compared methods with four metrics, where we have the following observations.

\begin{itemize}
    % phenomena
	\item \textcolor{black}{Our method achieves the best clustering results on most benchmark datasets.} For example, in the \textbf{non-graph} dataset Reuters, our approach improves the ARI, F1, ACC, and NMI values of SDCN \cite{bo2020structural} by 8.12\%, 3.33\%, 4.53\%, and 8.12\%, respectively. In the \textbf{graph} dataset DBLP, our approach improves 18.14\% over SDCN on ARI, 13.08\% on F1, 13.21\% on ACC, and 12.49\% on NMI. 
	% attention is good
	\item DAEGC enhances GAE by introducing the neighbor-wise embedding learning with an attention-based strategy, benefiting clustering performance improvement. Such a phenomenon validates the effectiveness of the attention-based mechanism. Differently, our method extends the attention-based mechanism to the heterogeneity-wise, scale-wise, and distribution-wise fusion modules to adaptively utilize the multiple off-the-shelf information, which significantly improves the clustering performance.
	% combinaiton is good
	\item SDCN performs better than the DAE-based (DAE, DEC, IDEC) and GCN-based (GAE, VGAE, ARGA) embedding clustering methods, demonstrating that combining DAE and GCN can contribute to clustering performance. Nevertheless, SDCN ($i$) equates the importance of the DAE feature and the GCN feature; ($ii$) neglects the multi-scale features; and ($iii$) fails to utilize available off-the-shelf information from the clustering assignment. \textcolor{black}{The proposed method addresses those issues and thus produces significantly better clustering performance than SDCN on all the datasets in almost all metrics. }
	% dual supervision is good
	\item Our method typically achieves better clustering performance than AGCN \cite{peng2021attention}, demonstrating the effectiveness of the proposed distribution-wise fusion module and the dual self-supervision solution in guiding the unsupervised clustering network training. For instance, in Amazon Photo, our approach improves 19.36\% on ARI, 28.00\% on F1, 20.22\% on ACC, and 14.51\% on NMI.
	% Why DBLP and Pubmed perfect work
	\item \textcolor{black}{Our method provides a significant improvement on DBLP and PubMed, e.g., in DBLP, our approach improves 10.29\% over the second-best one on ARI, 5.09\% on F1, 5.26\% on ACC, and 8.29\% on NMI. The possible reason is that DBLP and PubMed belong to datasets with low feature dimensions (i.e., little information), meaning that sufficiently utilizing the available off-the-shelf information plays a great important role in improving the clustering performance.}
	% why HHAR not works
	\item \textcolor{black}{Our method does not outperform AGCN in HHAR. The possible reason is that in HHAR, a series of dissimilar nodes are connected in the constructed KNN graph, reducing the graph quality. Although AGCN also uses the KNN graph, its auxiliary distribution $\mathbf{P}$ was inferred by the output of the conventional auto-encoder. Differently, the proposed method uses the graph convolutional network output to derive $\mathbf{P}$ for utilizing rich graph information of $\mathbf{Z}$. Thus, if the graph quality is terrible, the clustering performance of the proposed method may be worse than AGCN.}
	% Why AIDS
	\item \textcolor{black}{Our method obtains the best clustering performance on AIDS with four metrics, where AIDS is a large-scale long-tailed dataset that one class accounts for 62.34\% number, and the other thirty-seven classes share 37.66\%.}
\end{itemize}

\begin{figure}[]
	\centering
	\subfigure[$\mathbf{Q}$]{
	\includegraphics [width=0.21\columnwidth]{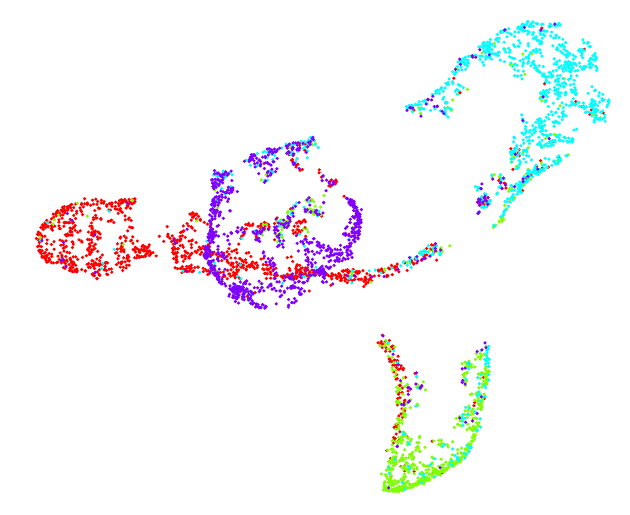}
	}
	\subfigure[$\mathbf{Z}$]{
	\includegraphics [width=0.21\columnwidth]{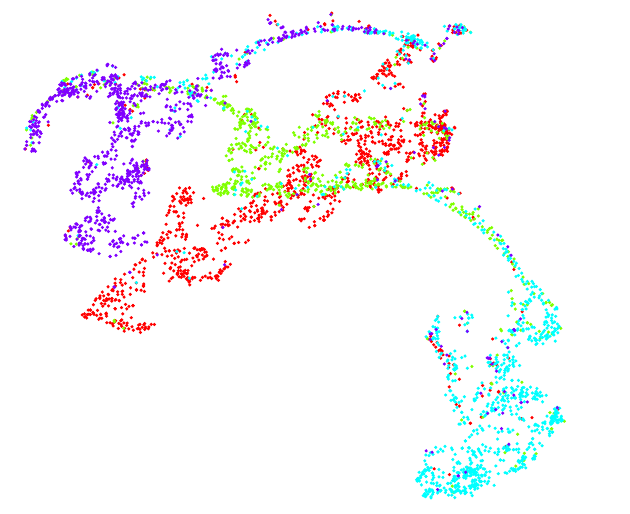}
	}
% 	\subfigure[0.5$\mathbf{Q}$+0.5$\mathbf{Z}$]{
% 	\includegraphics [width=0.21\columnwidth]{Figures/Our_Q3+Z3.pdf}
% 	}
	\subfigure[$\mathbf{F}$]{
	\includegraphics [width=0.23\columnwidth]{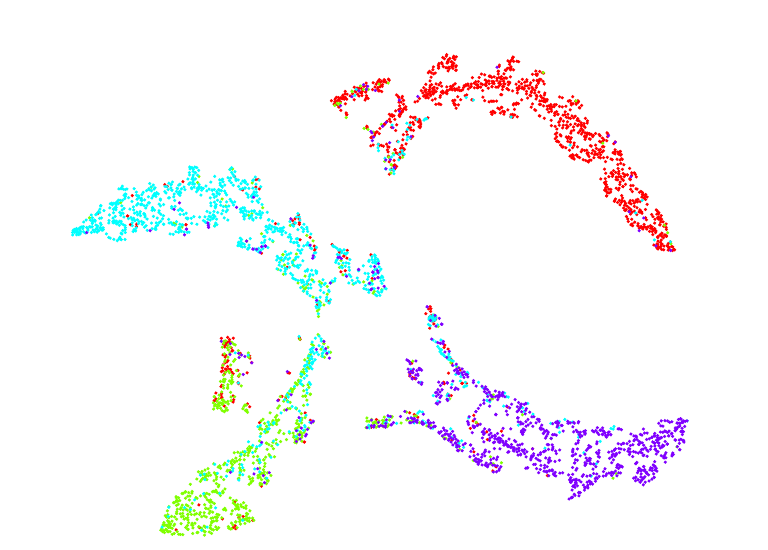}
	}
    \caption{The visual comparison of (a) distribution $\mathbf{Q}$, (b) distribution $\mathbf{Z}$, and (c) our adaptively fused one.}
	\label{fig: tsne1}
\end{figure}

\begin{figure}[]
	\centering
	\subfigure[USPS]{
	\includegraphics [width=0.29\columnwidth]{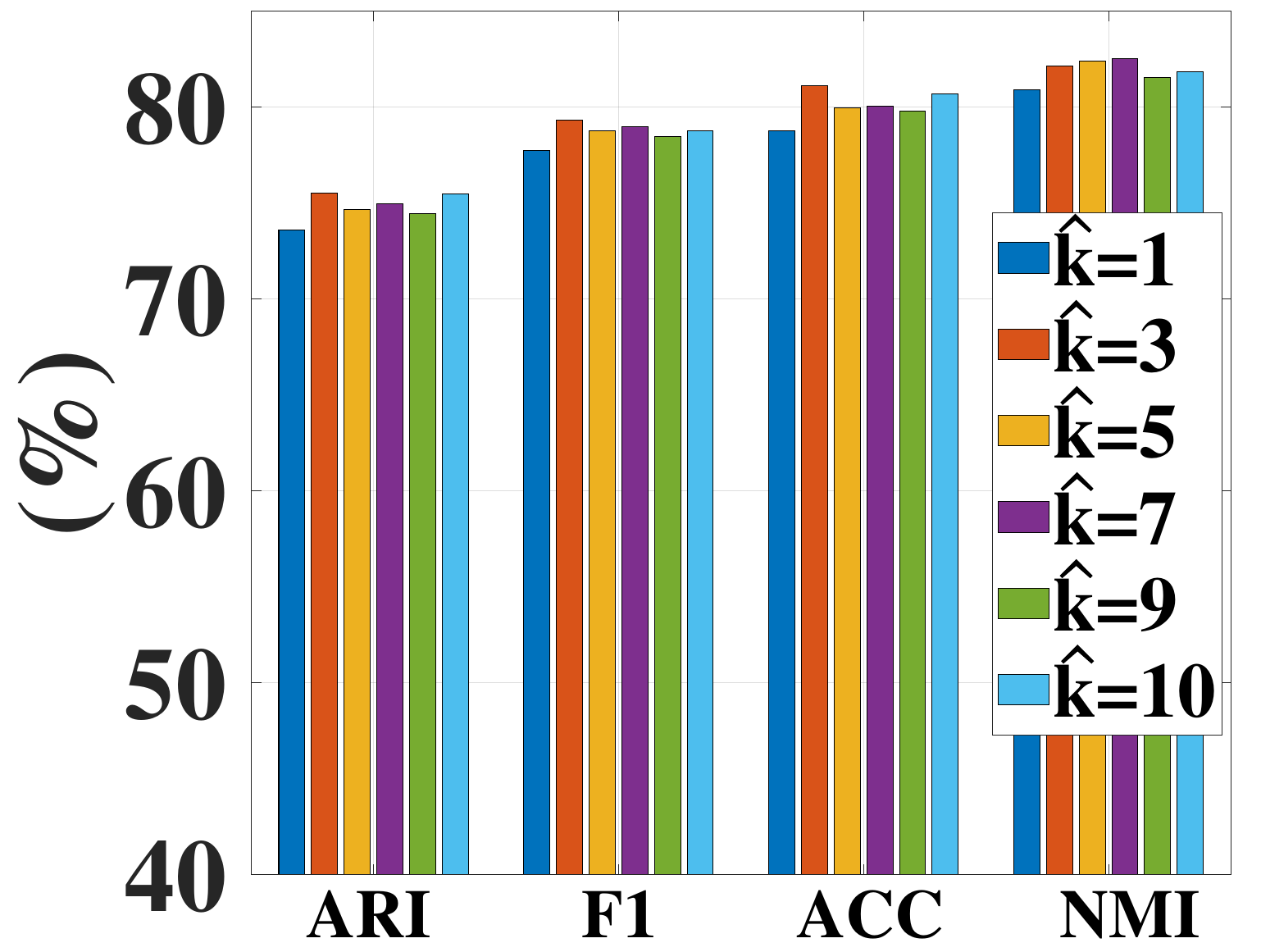}
	}
	\subfigure[Reuters]{
	\includegraphics [width=0.29\columnwidth]{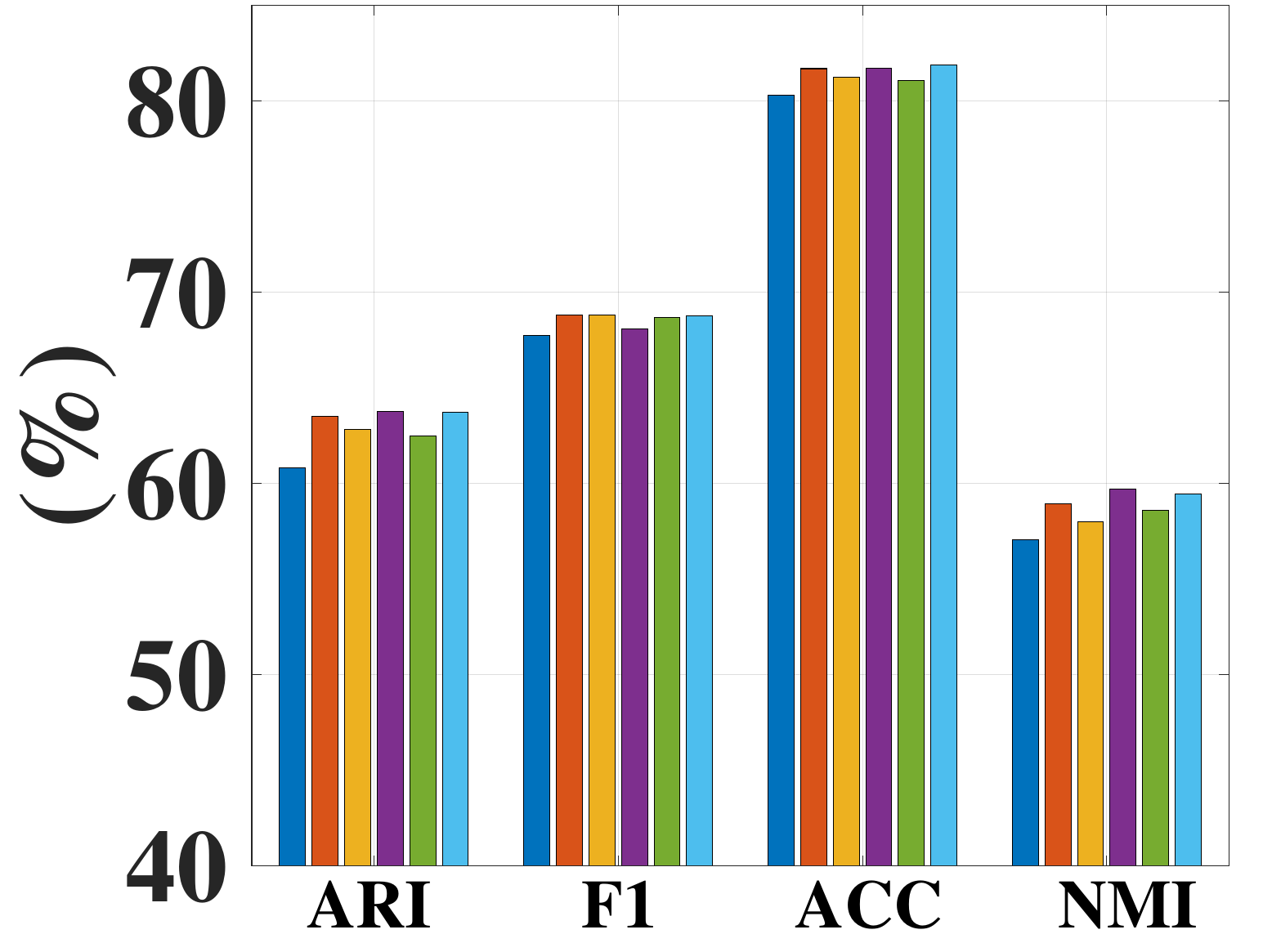}
	}
	\subfigure[HHAR]{
	\includegraphics [width=0.29\columnwidth]{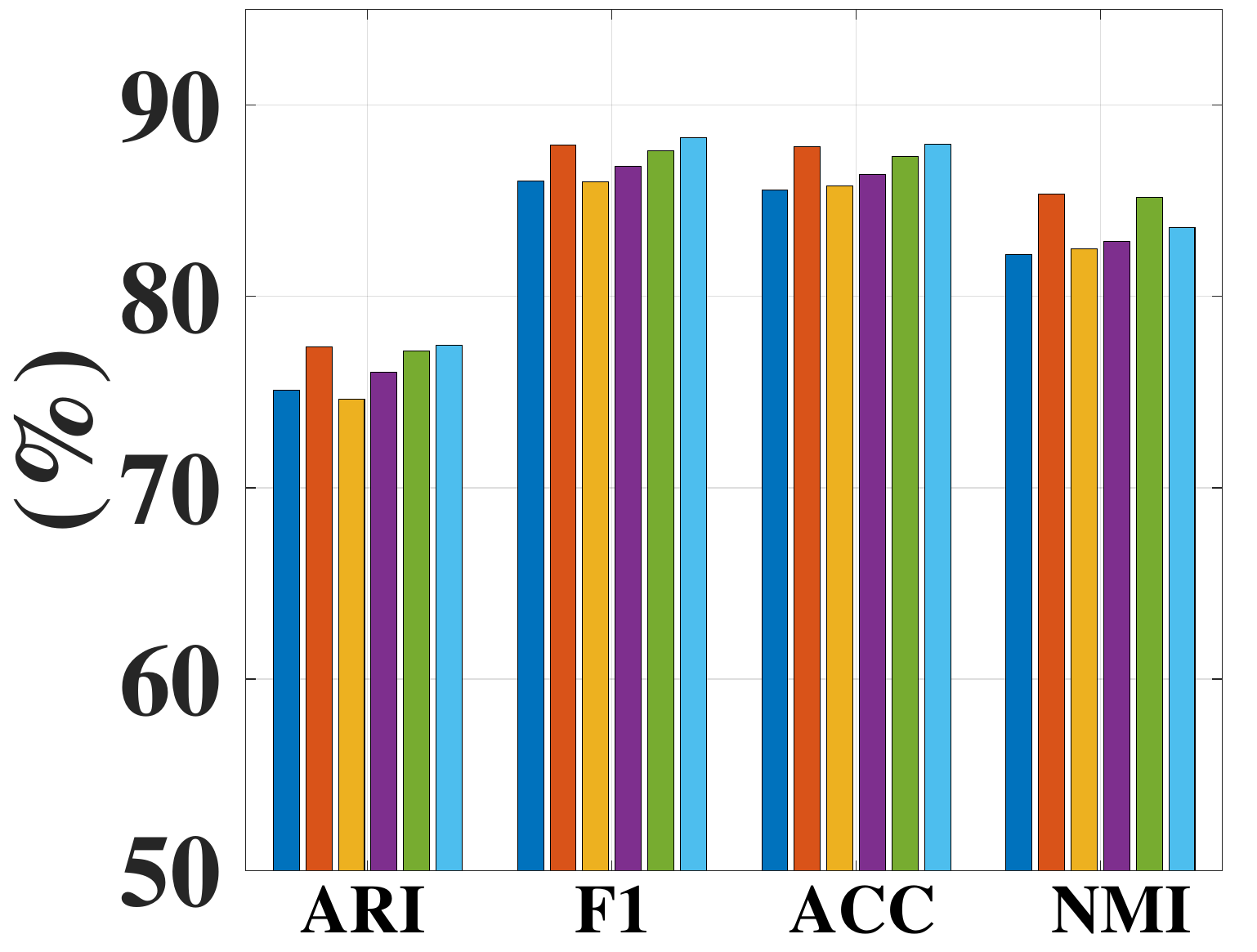}
	}
    \caption{Analyses of the number of neighbors for KNN graph construction. All the sub-figures share the same legend.}
	\label{fig: AS-k}
\end{figure}

\begin{figure*}[]
	\centering
	\subfigure[ARI]{
	\includegraphics [width=0.46\columnwidth]{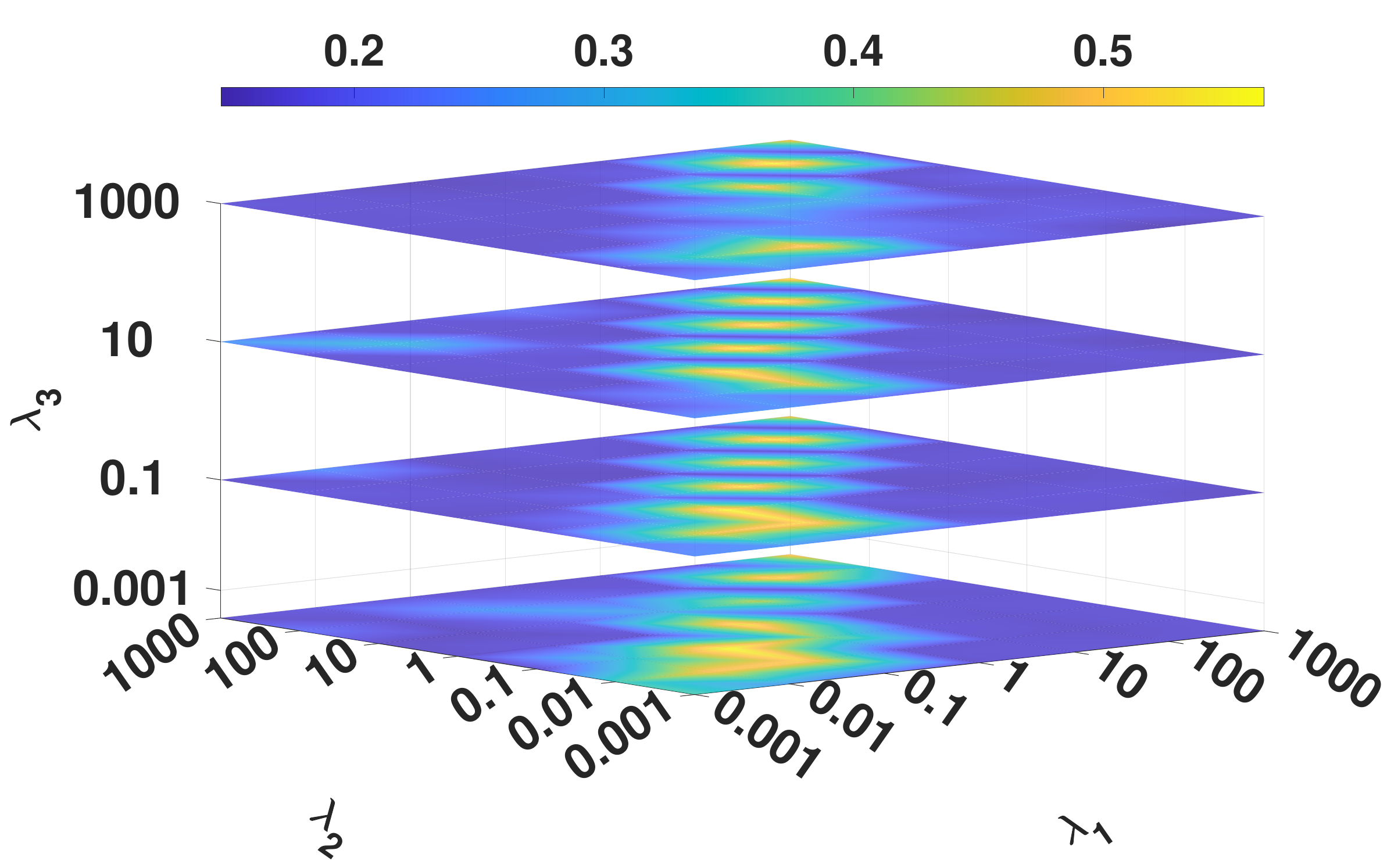}
	}
	\subfigure[F1]{
	\includegraphics [width=0.46\columnwidth]{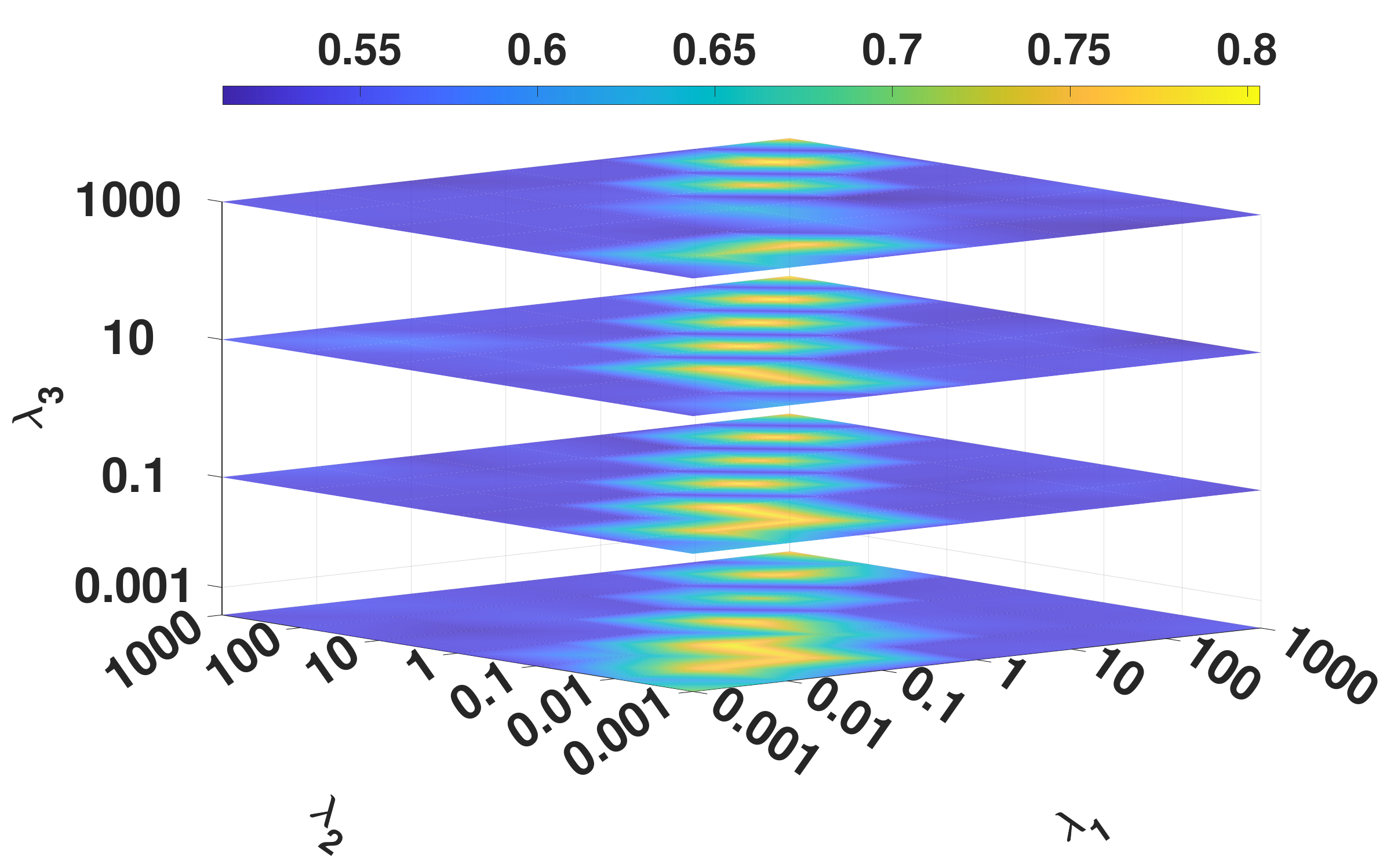}
	}
	\subfigure[ACC]{
	\includegraphics [width=0.46\columnwidth]{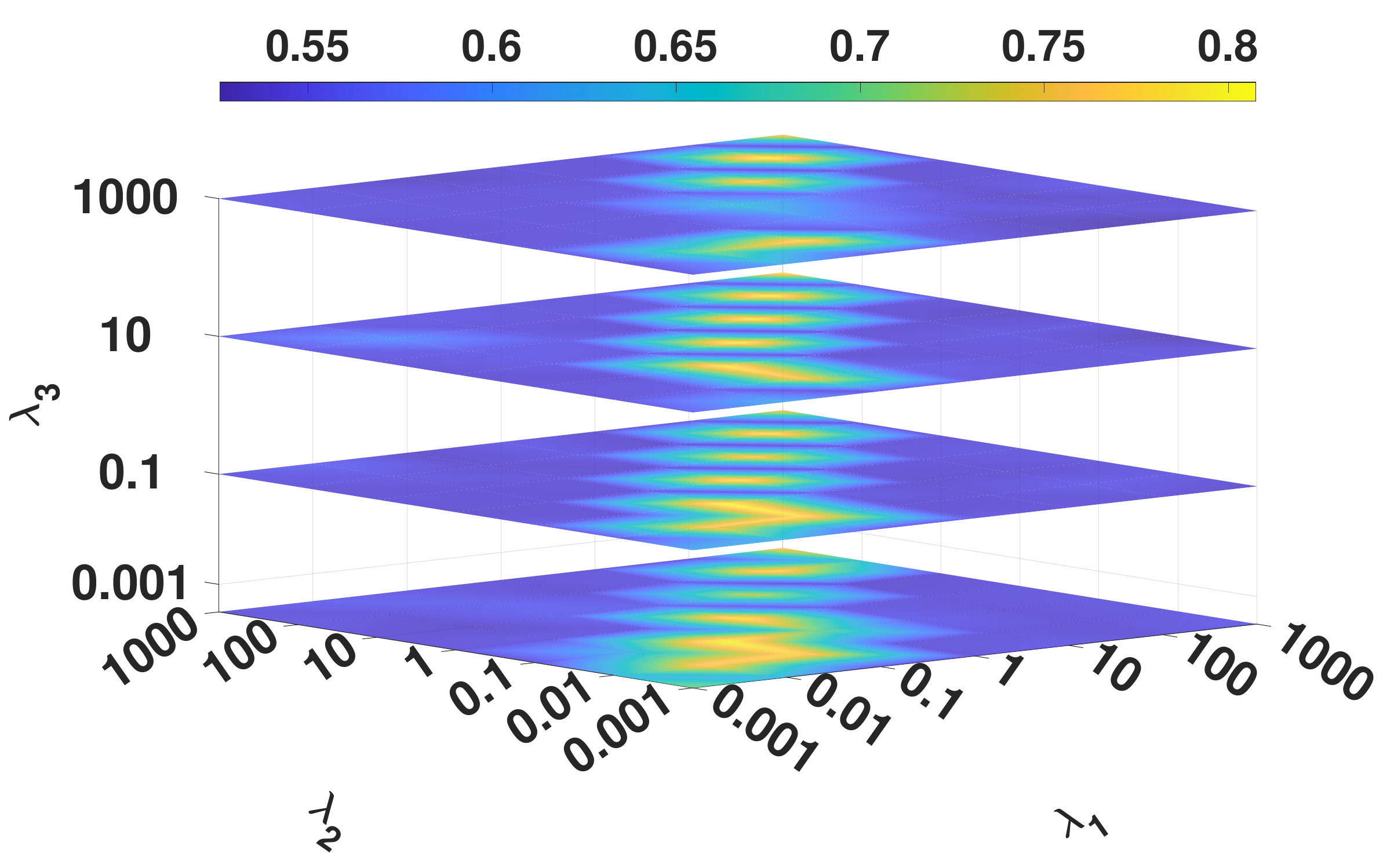}
	}
	\subfigure[NMI]{
	\includegraphics [width=0.46\columnwidth]{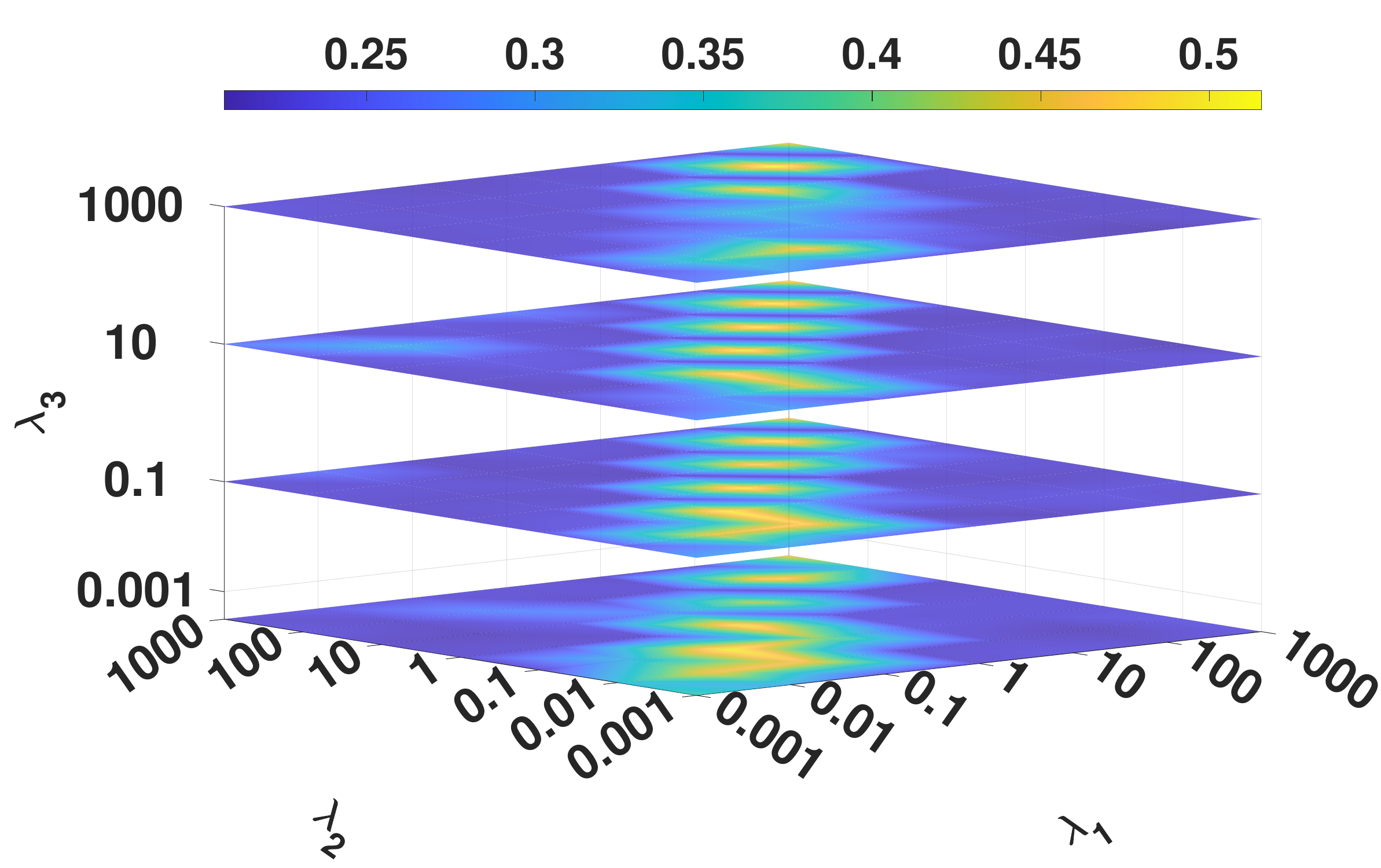}
	}
    \caption{Analysis of different hyperparameters ($\lambda_1$, $\lambda_2$, and $\lambda_3$) with four metrics on DBLP. We illustrate the results w.r.t. these hyperparameters in a 4D figure manner, where the color indicates the fourth direction, i.e., the corresponding experimental results.}
	\label{fig: PA_regs}
\end{figure*}

\begin{figure}[]
	\centering
	\subfigure[]    {\includegraphics [width=0.29\columnwidth]{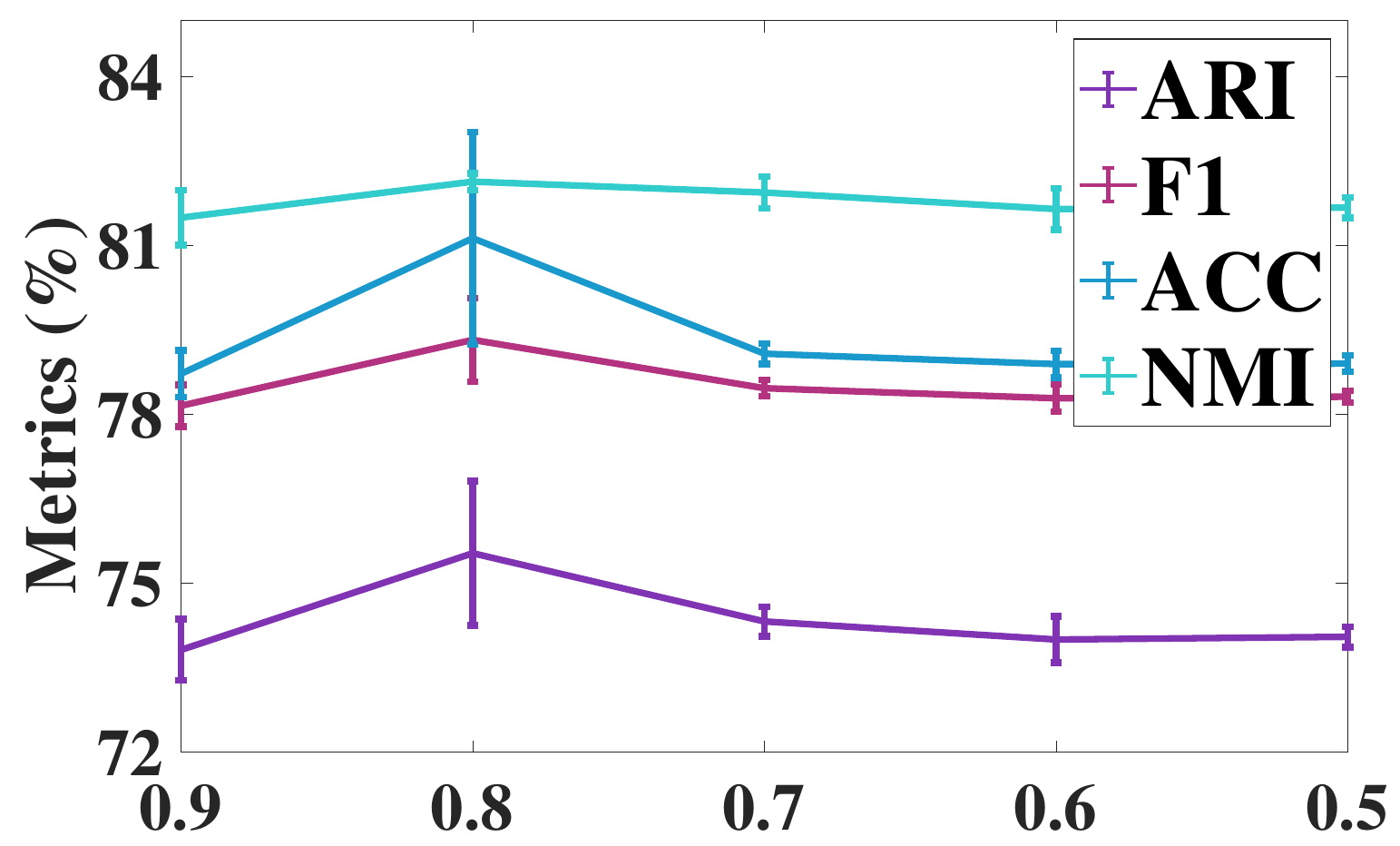}}
	\subfigure[] {\includegraphics [width=0.29\columnwidth]{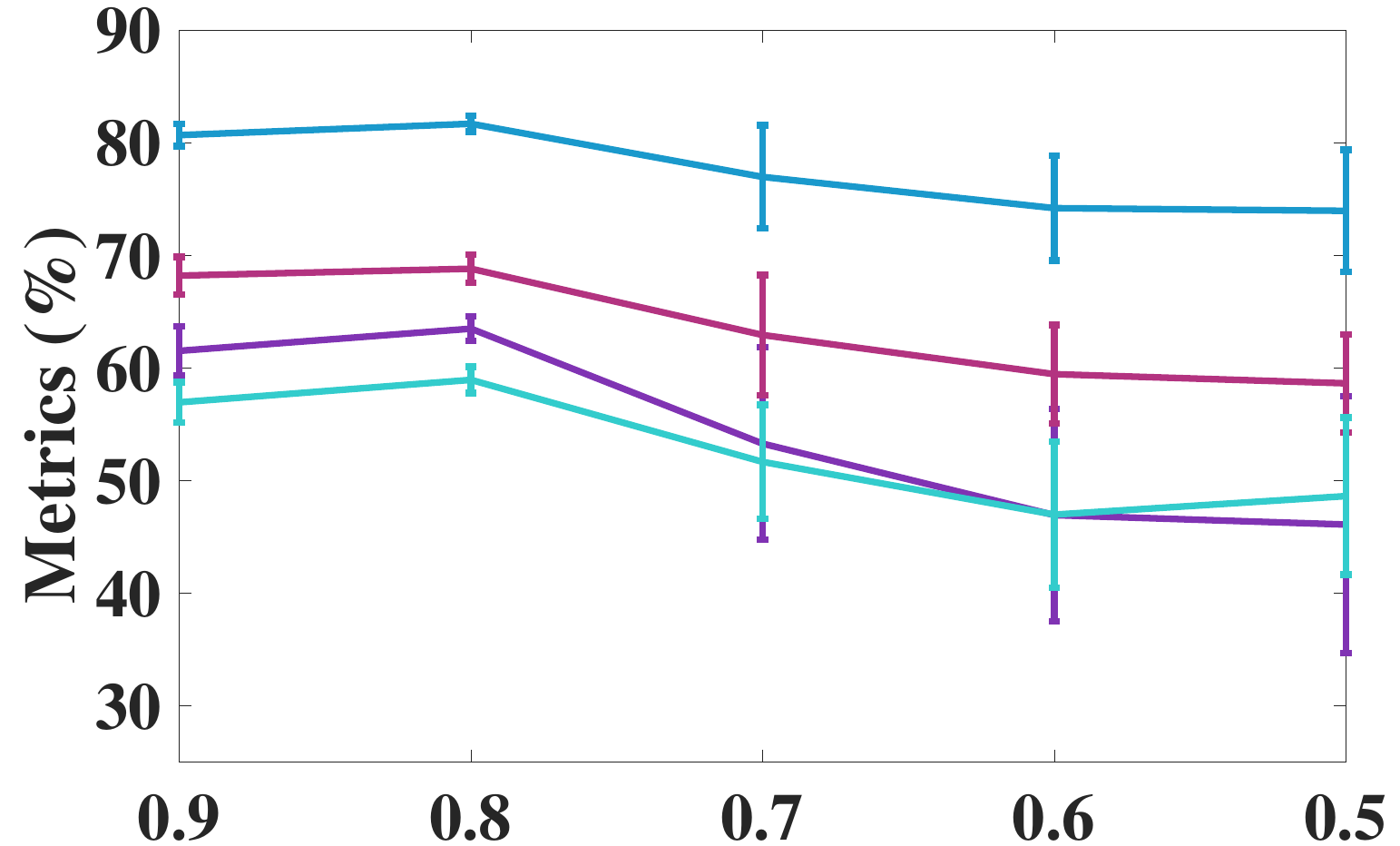}}
	\subfigure[]    {\includegraphics [width=0.29\columnwidth]{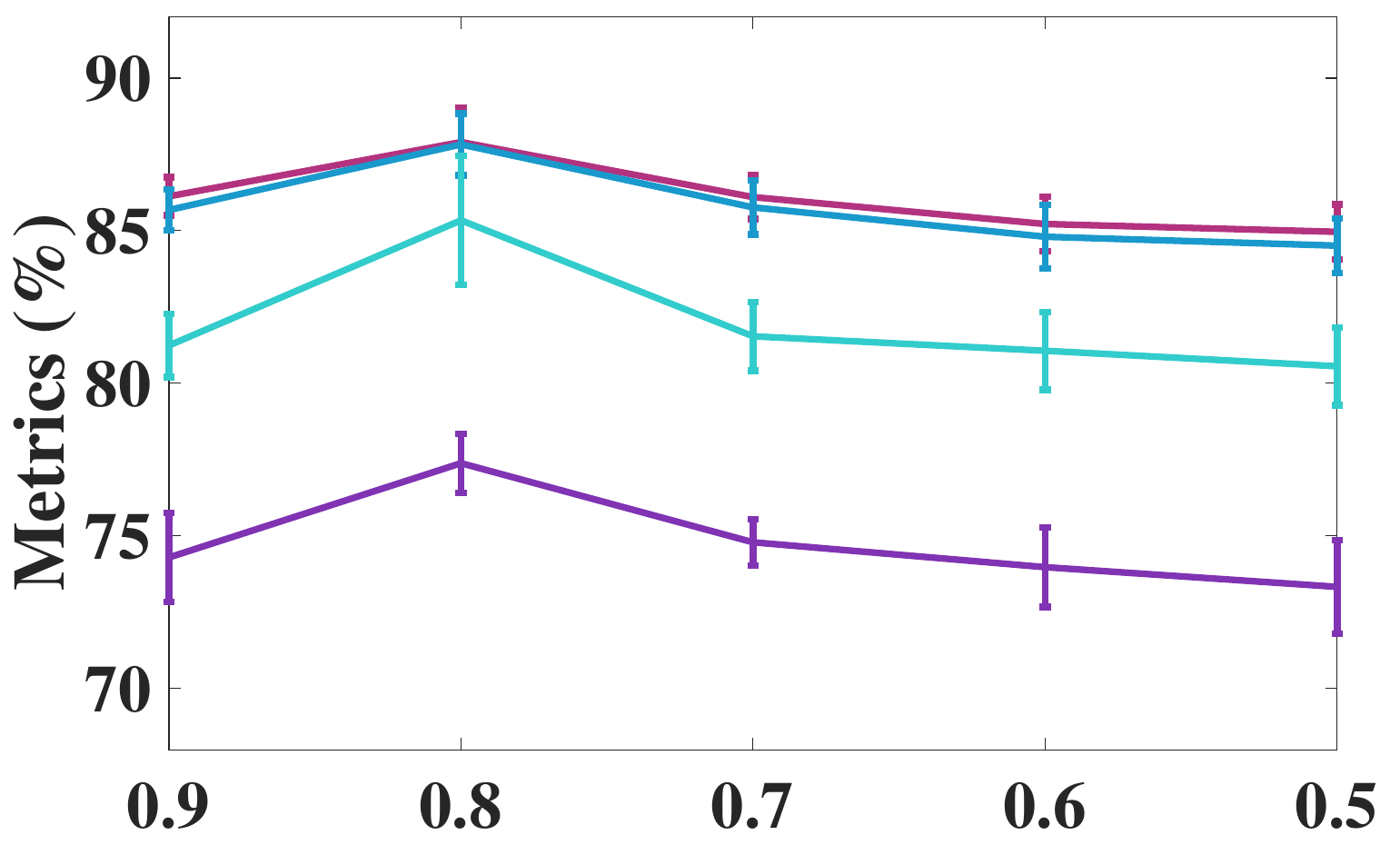}}
	\subfigure[]     {\includegraphics [width=0.29\columnwidth]{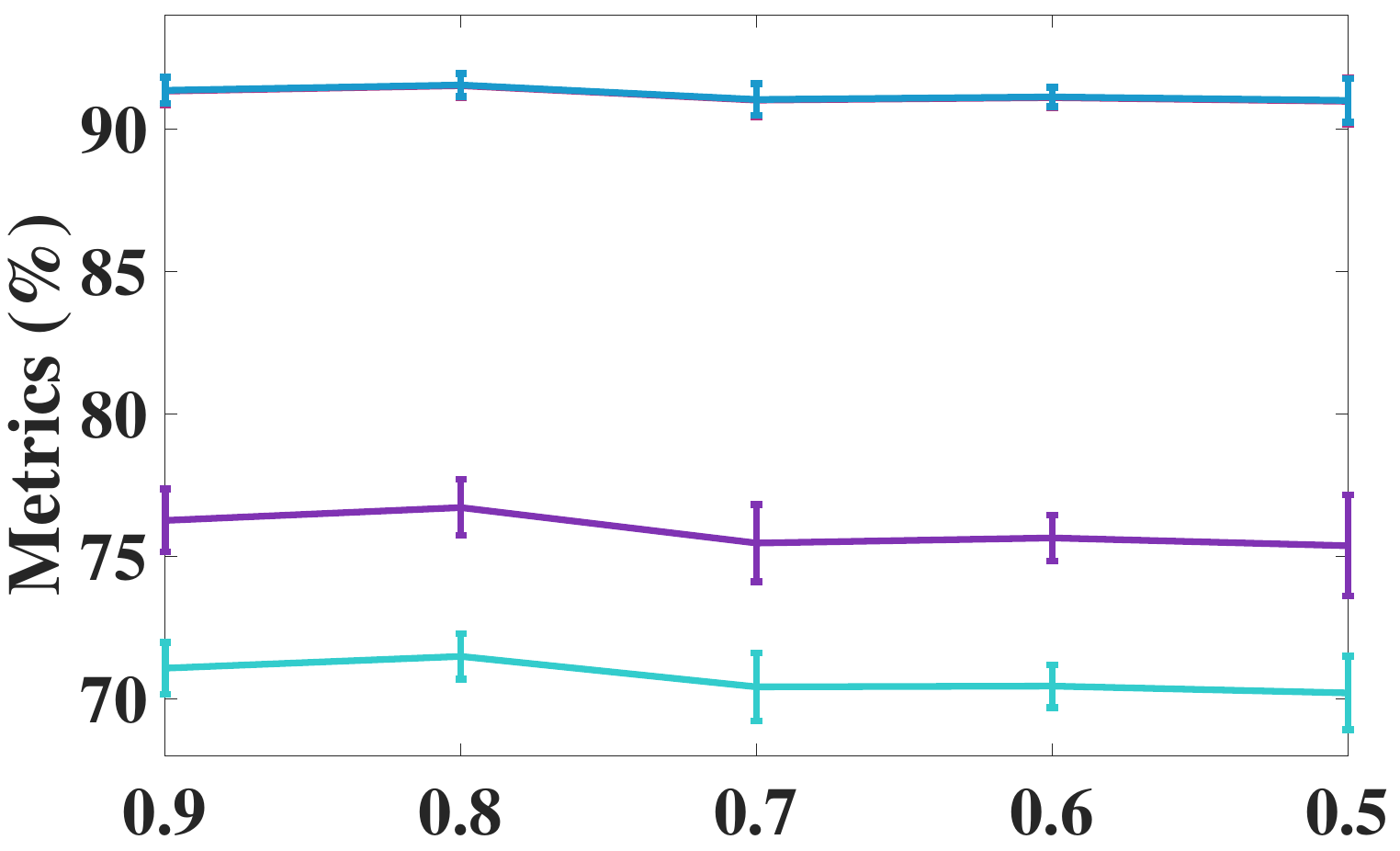}}
	\subfigure[]{\includegraphics [width=0.29\columnwidth]{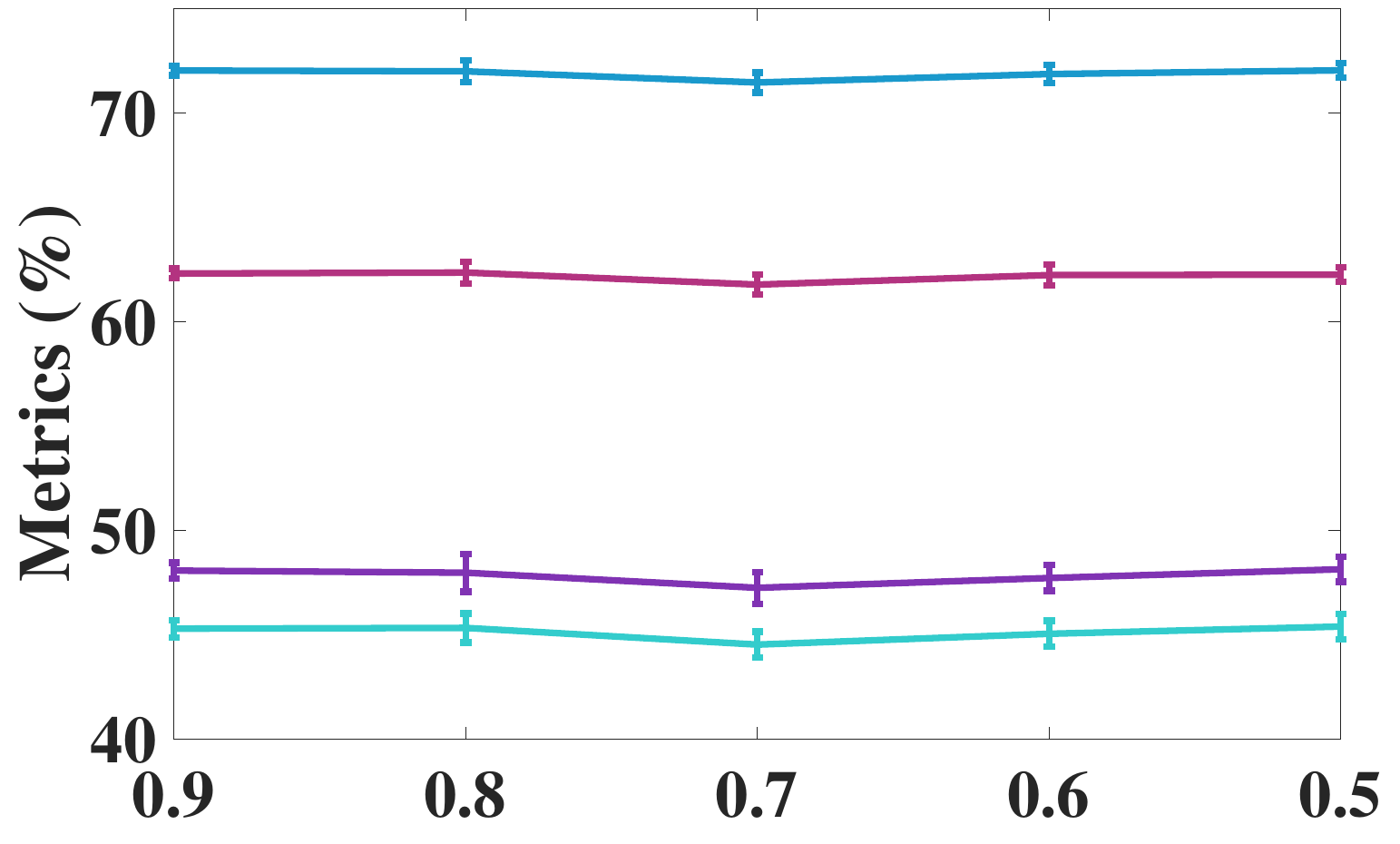}}
	\subfigure[]    {\includegraphics [width=0.29\columnwidth]{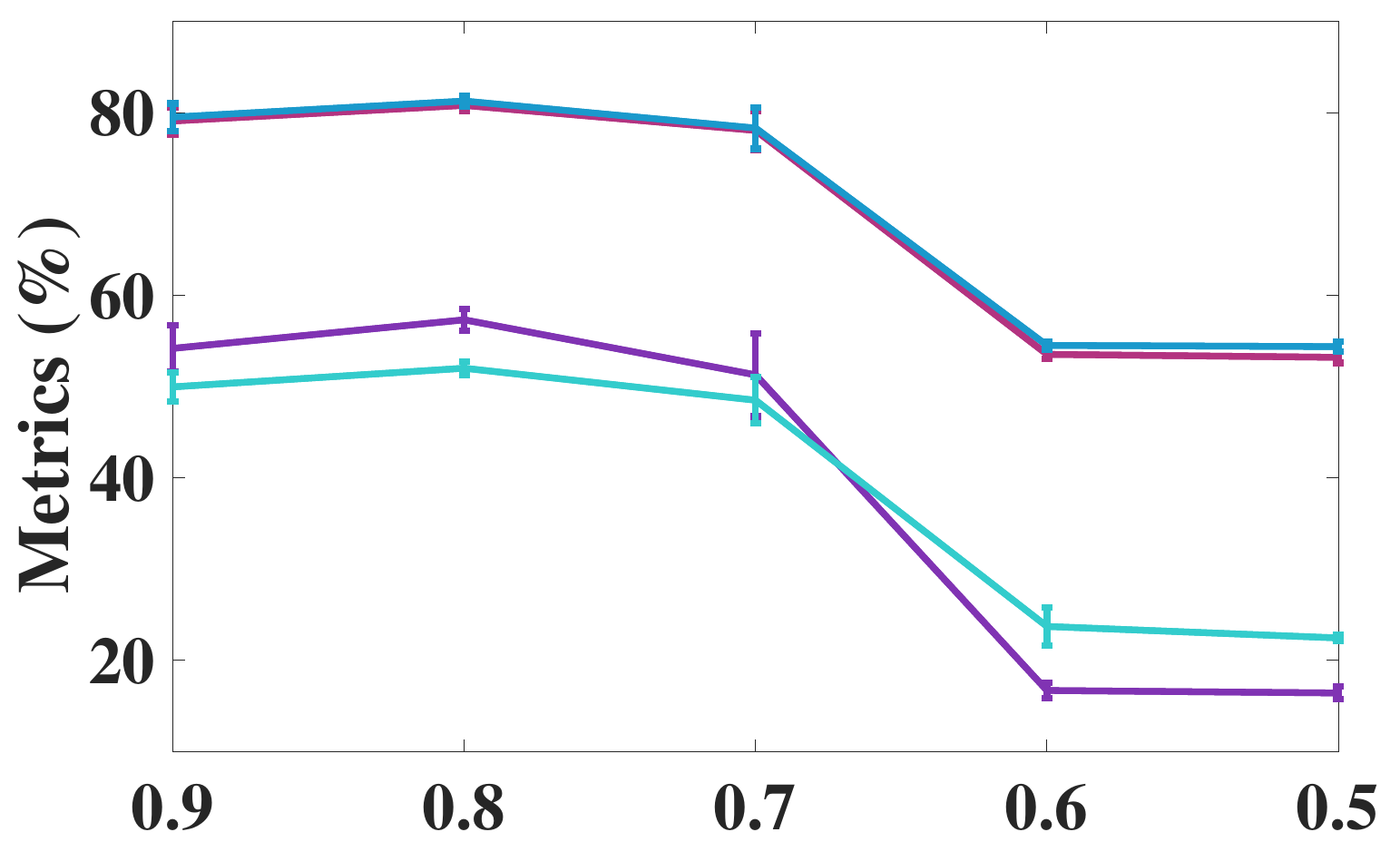}}
    \subfigure[]    {\includegraphics [width=0.29\columnwidth]{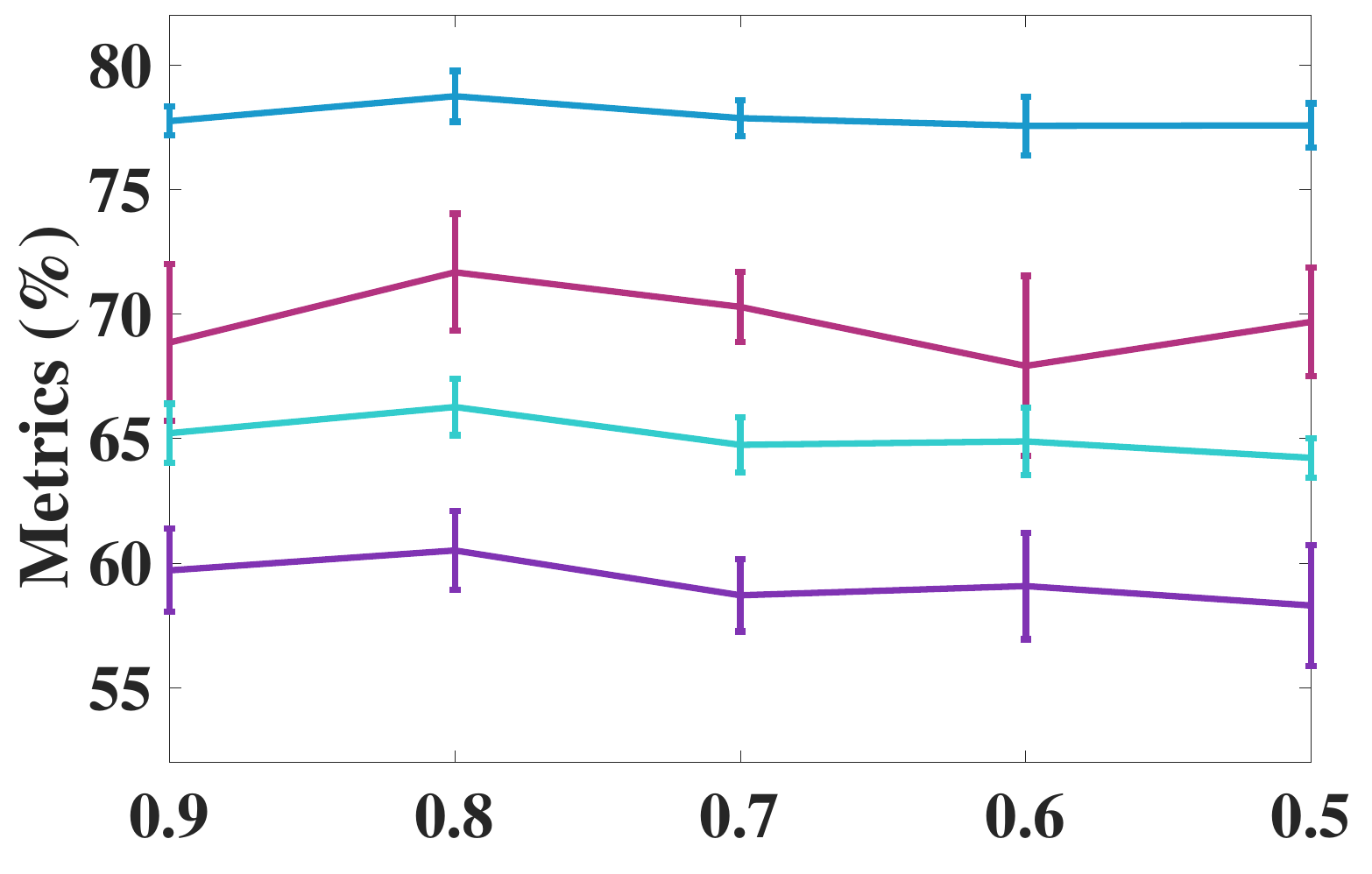}}
	\subfigure[]  {\includegraphics [width=0.29\columnwidth]{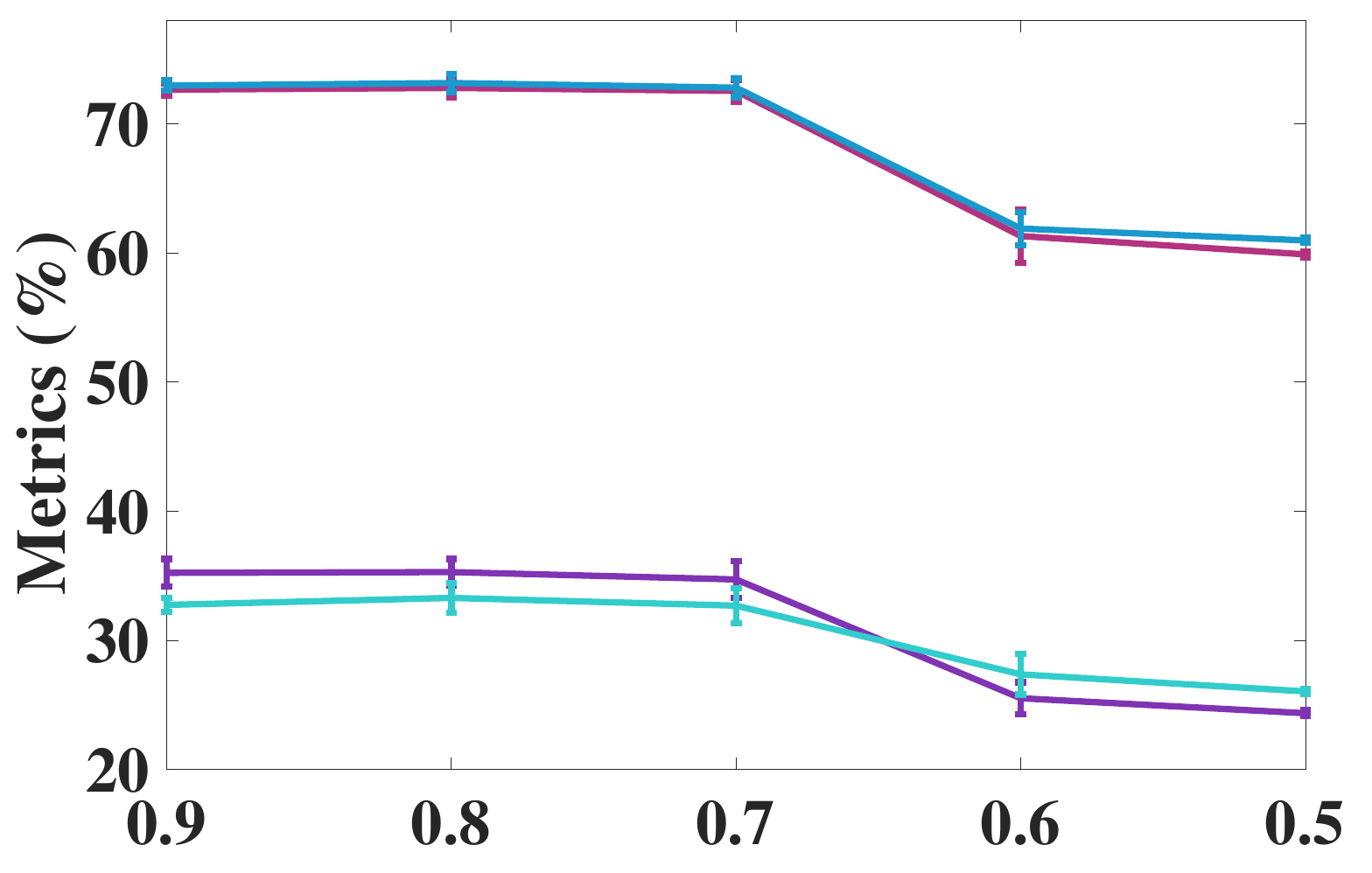}}
	\subfigure[]  {\includegraphics [width=0.30\columnwidth]{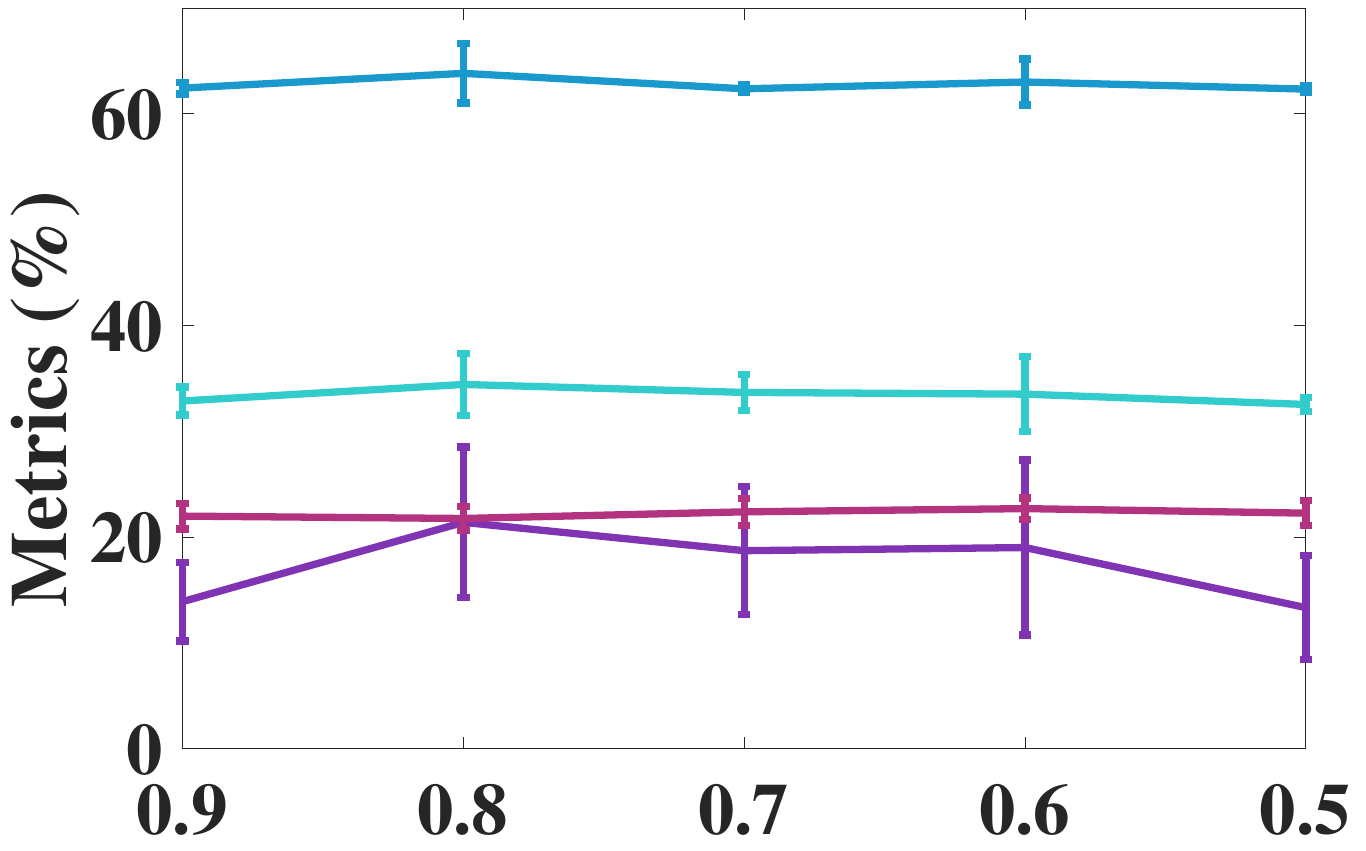}}
	\caption{Investigation of the effect of the threshold value for pseudo supervision on (a) USPS, (b) Reuters, (c) HHAR, (d) ACM, (e) CiteSeer, (f) DBLP, (g) Amazon Photo, (h) PubMed, and (i) \textcolor{black}{AIDS}.}
	\label{fig: thres}
\end{figure}

\subsection{Ablation Study}\label{sec: as}
We conducted comprehensive ablation studies to validate the effectiveness of the proposed modules and self-supervision strategies. \textcolor{black}{Table \ref{tab: AS} lists the quantitative results, where the first row of each dataset denotes the baseline that merges the DAE and GCN features in a half-and-half mechanism (i.e., without HWF) and uses the last GCN layer feature (i.e., without SWF and DWF) to conduct the optimization via the reconstruction loss and self-optimizing embedding loss following \cite{xie2016unsupervised,guo2017improved,wang2019attributed,bo2020structural,peng2021attention} (i.e., without SSS and HSS). The second, third, and fourth rows denote the baseline that adopts the proposed module HWF, HWF+SWF, and HWF+SWF+DWF, respectively. The fifth, sixth, and seventh rows denote the methods optimized by the introduced hard self-supervision, the soft self-supervision, and both.}

\textcolor{black}{
\subsubsection{Heterogeneity-wise fusion module}
By comparing the first and second rows of each dataset in Table \ref{tab: AS}, we can observe that HWF can typically improve clustering performance in most cases, validating its effectiveness. For example, on Reuters, it produces a 5.01\% performance improvement on ARI, 2.19\% on F1, 2.00\% on ACC, and 4.20\% on NMI.
}

\subsubsection{Scale-wise fusion module}
We can examine the effectiveness of SWF by comparing the second and third rows of each dataset in Table \ref{tab: AS}, in which the compared results with four metrics indicate the superiority of the SWF module in most datasets.

\subsubsection{Distribution-wise fusion module}
By comparing the results of the third and fourth rows of each dataset in Table \ref{tab: AS}, we observe that DWF also improves clustering performance, benefiting from the adaptive fusion of the information of two distributions.

\textcolor{black}{
To qualitatively validate the significant performance of the DWF module, we plotted 2D t-distributed stochastic neighbor embedding (t-SNE) \cite{maaten2008visualizing} visualizations of the distributions $\mathbf{Q}$, $\mathbf{Z}$, and $\mathbf{F}$ on DBLP in Figure \ref{fig: tsne1}, where we can see that our adaptively aggregated one is better than others, benefiting from adaptively (due to the DWF module) and effectively (due to the dual self-supervision solution) merging the information of two distributions.}

\textcolor{black}{
\subsubsection{Hard self-supervision strategy}
From the results of the fourth and fifth rows of each dataset in Table \ref{tab: AS}, it can be seen that on the \textbf{non-graph} dataset HHAR and \textbf{graph} dataset DBLP, there is about 2.00\% improvement when involving HSS, validating its effectiveness.
}

\subsubsection{Soft Self-supervision Strategy}
We can validate the effectiveness of SSS by comparing the results of the fourth and sixth rows of each dataset in Table \ref{tab: AS}. Specifically, on DBLP, SSS produces 13.73\% improvement on ARI, 7.15\% on F1, 7.37\% on ACC, and 10.82\% on NMI. Such impressive improvement is credited to that the SSS strategy refines the cluster assignment by minimizing a Kullback-Leibler divergence loss to promote consistent distribution alignment among distributions $\mathbf{Q}$, $\mathbf{Z}$, and $\mathbf{P}$.

\subsubsection{Dual Self-supervision (DSS)}
\textcolor{black}{By comparing the results of the fourth, fifth, sixth, and seventh rows of each dataset in Table \ref{tab: AS}, we can observe that DSS, which combines HSS and SSS, almost produces the best results on all nine benchmark datasets.
}

\begin{table*}[]
\centering
\caption{\textcolor{black}{The experiments on DBLP with the learned weights from the attention modules and four metrics results of corresponding representations.  The larger weight values are highlighted in \textbf{bold}, and the better clustering results are highlighted in \textcolor{red}{red}.}}
\label{tab: learn_weig}
\resizebox{0.86\textwidth}{!}{
\begin{tabular}{c||cc|cc|cc|ccccc|cc}
\hline\hline
$Sample$   &  $\mathbf{m}_{1,1}$ ($\mathbf{Z}_{1}$)   & $\mathbf{m}_{1,2}$ ($\mathbf{H}_{1}$)    &   $\mathbf{m}_{2,1}$ ($\mathbf{Z}_{2}$)   & $\mathbf{m}_{2,2}$ ($\mathbf{H}_{2}$)      & $\mathbf{m}_{3,1}$ ($\mathbf{Z}_{3}$)   & $\mathbf{m}_{3,2}$ ($\mathbf{H}_{3}$)  & $\mathbf{u}_{1}$ ($\mathbf{Z}_{1}$)    & $\mathbf{u}_{2}$ ($\mathbf{Z}_{2}$)    & $\mathbf{u}_{3}$ ($\mathbf{Z}_{3}$)    & $\mathbf{u}_{4}$ ($\mathbf{Z}_{4}$)    & $\mathbf{u}_{5}$ ($\mathbf{Z}_{5}$)    & $\mathbf{v}_{1}$ ($\mathbf{Z}$)    & $\mathbf{v}_{2}$ ($\mathbf{Q}$)      \\
\hline\hline
$\mathbf{x}_0$     & 0.6417          & \textbf{0.7669} & 0.0954       & \textbf{0.9954} & \textbf{0.7091}       & 0.7051          & 0.3874 & 0.3896 & 0.4718          & 0.3930       & \textbf{0.5666} & \textbf{0.8980} & 0.4399       \\
$\mathbf{x}_1$     & 0.6166          & \textbf{0.7872} & 0.0918       & \textbf{0.9958} & 0.7052       & \textbf{0.7090} & 0.2780 & 0.2783 & 0.2854          & 0.2824       & \textbf{0.8271} & \textbf{0.8584} & 0.5131       \\
$\mathbf{x}_2$     & \textbf{0.8232} & 0.5678          & 0.2532       & \textbf{0.9674} & 0.7051       & \textbf{0.7091} & 0.2188 & 0.2187 & \textbf{0.8826} & 0.2216       & 0.2762          & \textbf{0.8585} & 0.5128       \\
\vdots & \vdots           & \vdots           & \vdots        & \vdots           & \vdots        & \vdots           & \vdots  & \vdots  & \vdots           & \vdots        & \vdots           & \vdots           & \vdots        \\
$\mathbf{x}_{4054}$  & 0.6254          & \textbf{0.7803} & 0.0686       & \textbf{0.9976} & 0.7047       & \textbf{0.7095} & 0.2503 & 0.2505 & 0.2572          & 0.2544       & \textbf{0.8624} & \textbf{0.8582} & 0.5134       \\
$\mathbf{x}_{4055}$  & 0.6238          & \textbf{0.7816} & 0.0508       & \textbf{0.9987} & 0.7060       & \textbf{0.7082} & 0.2733 & 0.2741 & 0.2817          & 0.2782       & \textbf{0.8327} & \textbf{0.8584} & 0.5129       \\
$\mathbf{x}_{4056}$  & 0.5919          & \textbf{0.8060} & 0.0811       & \textbf{0.9967} & \textbf{0.7082}       & 0.7060          & 0.3903 & 0.3926 & \textbf{0.5544} & 0.3956       & 0.4793          & \textbf{0.8979} & 0.4402       \\
\hline
AVG   & \textbf{0.7182} & 0.6704          & 0.1310       & \textbf{0.9854} & 0.5916       & \textbf{0.7698} & 0.3162 & 0.3157 & 0.5141 & 0.3200       & \textbf{0.5821} & \textbf{0.8555} & 0.5138       \\
\hline\hline
ARI   & \textcolor{black}{0.0836}    & \textcolor{red}{0.5539} & \textcolor{black}{0.3009} & \textcolor{red}{0.5500} & \textcolor{black}{0.5437} & \textcolor{red}{0.5489} & 0.0836 & 0.3009 & 0.5437 & \textcolor{black}{0.5453} & \textcolor{red}{0.5489} & \textcolor{red}{0.5500} & \textcolor{black}{0.5489} \\
F1    & \textcolor{black}{0.4072}    & \textcolor{red}{0.7972} & \textcolor{black}{0.5781} & \textcolor{red}{0.7969} & \textcolor{black}{0.7941} & \textcolor{red}{0.7964} & 0.4072 & 0.5781 & 0.7941 & \textcolor{black}{0.7944} & \textcolor{red}{0.7964} & \textcolor{red}{0.7969} & \textcolor{black}{0.7964} \\
ACC   & \textcolor{black}{0.4212}    & \textcolor{red}{0.8023} & \textcolor{black}{0.6120} & \textcolor{red}{0.8011} & \textcolor{black}{0.7981} & \textcolor{red}{0.8006} & 0.4212 & 0.6120 & 0.7981 & \textcolor{black}{0.7986} & \textcolor{red}{0.8006} & \textcolor{red}{0.8011} & \textcolor{black}{0.8006} \\
NMI   & \textcolor{black}{0.1778}    & \textcolor{red}{0.4987} & \textcolor{black}{0.3023} & \textcolor{red}{0.4996} & \textcolor{black}{0.4950} & \textcolor{red}{0.4988} & 0.1778 & 0.3023 & 0.4950 & \textcolor{black}{0.4965} & \textcolor{red}{0.4988} & \textcolor{red}{0.4995} & \textcolor{black}{0.4988} \\
\hline\hline
\end{tabular}
}
\end{table*}

\begin{table}
\centering
\caption{\textcolor{black}{Comparisons of the clustering results, network parameters, and spending time of compared methods on DBLP. The best results are highlighted in \textbf{bold}.}}
\label{tab: time}
\resizebox{0.48\textwidth}{!}{
\begin{tabular}{l|cccc|c}
\hline\hline
Metrics  & SDCN              & AGCN           & AGCC                        & Our                     & boost $\uparrow$  \\
\hline\hline
ARI (\%) & 39.15$\pm$2.01    & 42.49$\pm$0.31 & 44.40$\pm$3.79 & \textbf{57.29$\pm$1.20} & $\uparrow$ 12.89  \\
F1 (\%)  & 67.71$\pm$1.51    & 72.80$\pm$0.56 & 71.84$\pm$2.02 & \textbf{80.79$\pm$0.61} & $\uparrow$ 07.99  \\
ACC (\%) & 68.05$\pm$1.81    & 73.26$\pm$0.37 & 73.45$\pm$2.16 & \textbf{81.26$\pm$0.62} & $\uparrow$ 07.81  \\
NMI (\%) & 39.50$\pm$1.34    & 39.68$\pm$0.42 & 40.36$\pm$2.81 & \textbf{51.99$\pm$0.76} & $\uparrow$ 11.63  \\
\hline
Parameters (M) 
& \textbf{4.31742}%4.317424} 
& 4.35658%4.356575       
& 11.86304%11.863038                           
& 4.35659%4.356593                
&                  \\
Time (s) & \textbf{253.3905} & 273.1204       & 5420.5255                            & 310.6794                &                  \\
\hline\hline
\end{tabular}
}
\end{table}

\begin{figure}[]
	\centering
	\subfigure[]{
	\includegraphics [width=0.20\columnwidth]{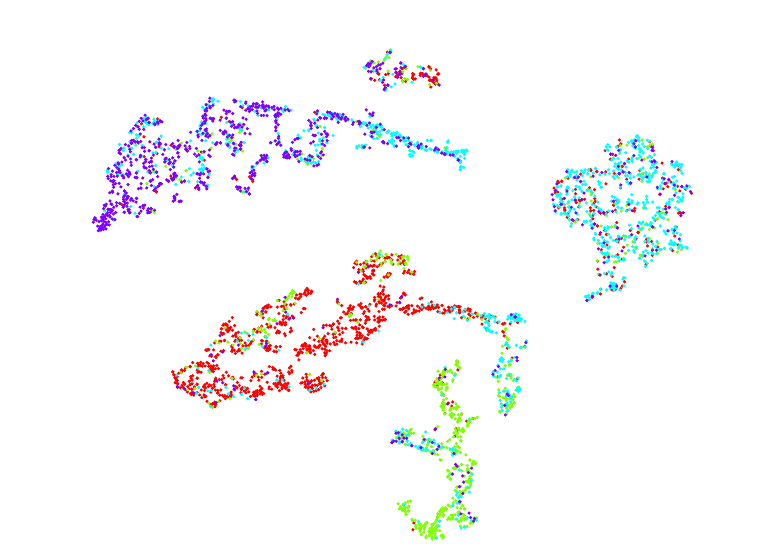}
	}
	\subfigure[]{
	\includegraphics [width=0.20\columnwidth]{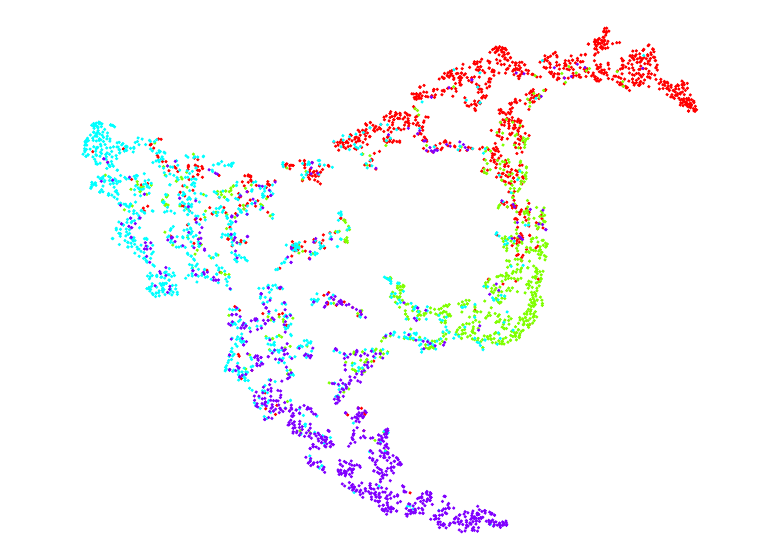}
	}
	\subfigure[]{
	\includegraphics [width=0.20\columnwidth]{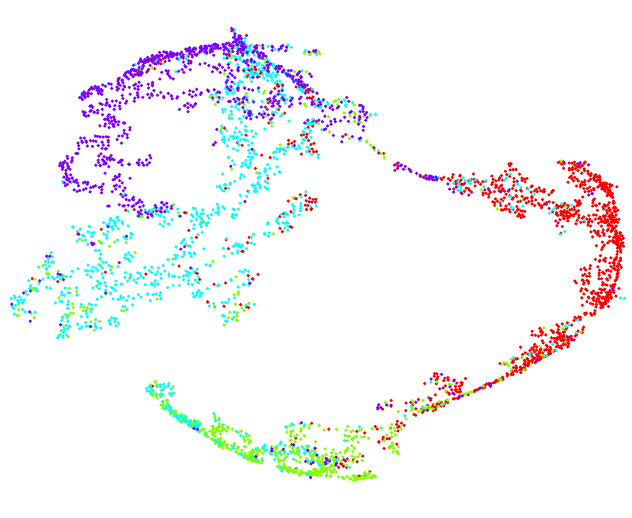}
	}
	\subfigure[]{
	\includegraphics [width=0.23\columnwidth]{Figures/Ours_F.pdf}
	}
    \caption{Comparison of the visualization of the learned representations on DBLP by (a) SDCN \cite{bo2020structural}, (b) AGCN \cite{peng2021attention}, (c) DFCN [\cite{tu2021deep}, and (d) Ours, where different colors represent different clusters.}
	\label{fig: tsne2}
\end{figure}

\subsection{Parameters Analysis}

\subsubsection{Analysis of the number of neighbors}
As the number of neighbors $\hat{\emph{k}}$ directly decides the KNN graph with respect to (w.r.t.) the quality of the adjacency matrix, we tested different $\hat{\emph{k}}$ on the non-graph datasets, i.e., USPS, Reuters, and HHAR. From Figure \ref{fig: AS-k}, we can observe that our model is not sensitive to $\hat{\emph{k}}$. In the experiments, we fixed $\hat{\emph{k}}$ to $3$ to construct the KNN graph for the non-graph datasets.

\subsubsection{Analysis of hyperparameters}
We investigated the influence of the hyperparameters, i.e., $\lambda_1$, $\lambda_2$, and $\lambda_3$, on DBLP. Figure \ref{fig: PA_regs} illustrates four metrics results in a 4D figure manner where the color indicates the fourth direction, i.e., the corresponding experimental results. From Figure \ref{fig: PA_regs}, we have the following observations.
\begin{itemize}
    \item The parameters setting of $\lambda_1$ and $\lambda_2$ is critical to the proposed model. Specifically, the highest clustering result occurs when $\lambda_1$ and $\lambda_2$ tend to the same value. This phenomenon reflects the importance of balancing the regularization term in constraining the distribution alignment. 
    \item Our model is robust to the hyperparameter $\lambda_3$, i.e., our method can obtain the optimal performance in a wide and common parameter range of $\lambda_3$.
\end{itemize}

\subsubsection{Analysis of the threshold value}
We investigated the effect of the threshold value $r$ on clustering performance. Figure \ref{fig: thres} shows the clustering results with various thresholds (i.e., $0.5$, $0.6$, $0.7$, $0.8$, and $0.9$). From Figure \ref{fig: thres}, we have the following conclusions.
\begin{itemize}
    \item A small threshold value unavoidably degrades the clustering performance compared with the ones using a large threshold value. For example, when we set $r$ to $0.5$ or $0.6$, all four metrics results on DBLP have degraded performance. Apparently, a small threshold value can easily generate a lot of incorrect pseudo-labels.
    \item A large threshold value is capable of leading to high clustering performance. However, setting $r$ to a tremendous value like $0.9$ cannot improve clustering performance. The reason is that with a larger threshold, the number of selected supervised labels will reduce, resulting in weak label propagation. Thus, we set $r$ to $0.8$ in this paper.
\end{itemize}

\subsubsection{Analysis of the learned attention-aware weights}
\textcolor{black}{We added the results of the learned weight on DBLP to verify the effectiveness of the designed attention mechanism in Table \ref{tab: learn_weig}, \textcolor{black}{where $\mathbf{x}_j$ indicates the $\emph{j}$-th sample; $\mathbf{m}_{i,1}$ and $\mathbf{m}_{i,2}$ indicate the HWF learned weights of $\mathbf{Z}_{i}$ and $\mathbf{H}_{i}$ in the $\emph{i}$-th layer, respectively; $\mathbf{u}_{1}$, $\mathbf{u}_{2}$, $\mathbf{u}_{3}$, $\mathbf{u}_{4}$, and $\mathbf{u}_{5}$ indicate the SWF learned weights; $\mathbf{v}_{1}$ and $\mathbf{v}_{2}$ indicate the DWF learned weights of $\mathbf{Z}$ and $\mathbf{Q}$, respectively; AVG indicates the average value of the weight results. The clustering results of $\mathbf{Z}$ and $\mathbf{Q}$ are inferred through their column indexes of the maximum in each row, and those results of other features are obtained with K-means, where the higher clustering performance, the better the feature representation.} We can see that the representation corresponding to a large weight value typically performs better clustering results than the one corresponding to a small weight value, substantiating the effectiveness of the designed attention mechanism in the weighted fusion.}

\subsection{Time and Space Complexity Analysis}
\textcolor{black}{We repeated the experiment 10 times to compare the mean values, the standard deviations (i.e., mean$\pm$std), the parameters number, and the running time of the proposed method with the baselines \cite{bo2020structural,peng2021attention,tu2021deep} on DBLP in Table \ref{tab: time}. Specifically, the experiments are implemented with Python 3.6.12 and  Pytorch-1.9.0+cu102 on an NVIDIA GeForce RTX 2080 Ti and an i7-8700K CPU. M and s are the abbreviations of the million and second, respectively.} From Table \ref{tab: time}, we can observe that our method obtains a significant clustering improvement at the cost of acceptable resource consumption. 

\subsection{Visual Comparison}
\textcolor{black}{To qualitatively evaluate the effectiveness of the proposed method, we plotted 2D t-SNE visualizations of baselines \cite{bo2020structural,peng2021attention,tu2021deep} and the proposed method on DBLP in Figure \ref{fig: tsne2}, where we can find that the feature representation obtained by our method shows the best separability for different clusters, i.e., samples from the same class naturally gather together and the gap between different groups is the most obvious one. This phenomenon substantiates that our method produces the most clustering-oriented representation compared with state-of-the-art methods.
}

\section{Conclusion}\label{sec: con}
We have presented a novel deep embedding clustering method that simultaneously enhances embedding learning and cluster assignment. Specifically, we first designed heterogeneity-wise and scale-wise fusion modules to learn an informative representation adaptively. Then, we utilized a distribution-wise fusion module to achieve cluster enhancement via an attention-based mechanism. Finally, we proposed a soft self-supervision strategy with a Kullback-Leibler divergence loss and a hard self-supervision strategy with a pseudo supervision loss to utilize the available off-the-shelf information from the cluster assignments. The quantitative and qualitative experiments and analyses demonstrate that our method consistently outperforms state-of-the-art approaches. We also provided comprehensive ablation studies to validate the effectiveness and advantage of our network.

\balance
\bibliographystyle{IEEEtran}
\bibliography{DGC}

\begin{IEEEbiography}[{\includegraphics[width=1in,height=1.75in,clip,keepaspectratio]{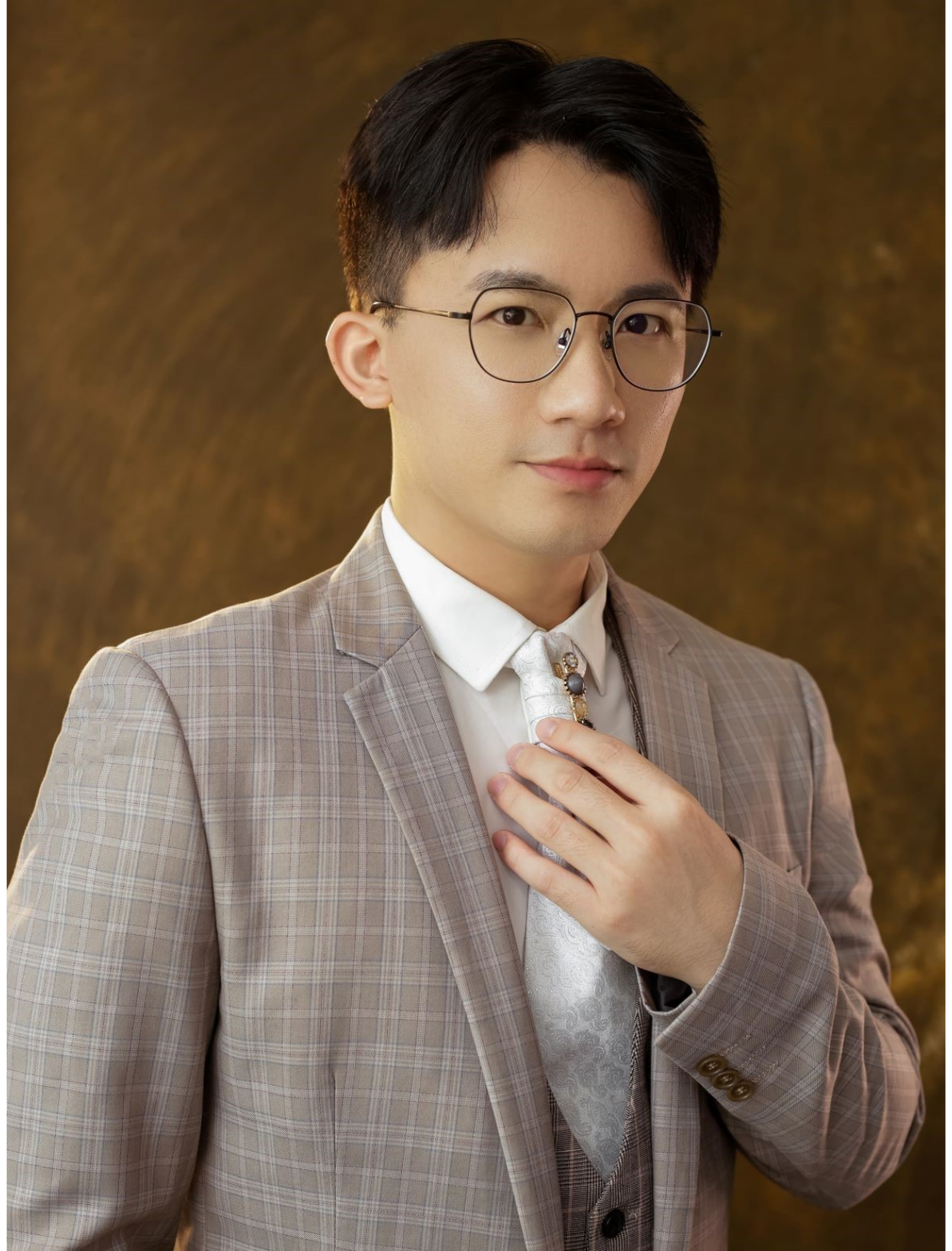}}]{Zhihao Peng}
received the B.S. and M.S. degrees in computer science and technology from Guangdong University of Technology, Guangzhou, China, in 2016 and 2019, respectively. He is currently pursuing the Ph.D. degree in department of computer science from City University of Hong Kong, SAR, China. His current research interests include spectral clustering, subspace learning, and domain adaptation in image/text/graph processing with unsupervised learning.
\end{IEEEbiography}

\vspace{-10 mm}

\begin{IEEEbiography}[{\includegraphics[width=1in,height=1.75in,clip,keepaspectratio]{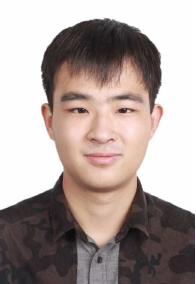}}]{Yuheng Jia}
received the B.S. degree in automation and the M.S. degree in control theory and engineering from Zhengzhou University, Zhengzhou, China, in 2012 and 2015, respectively, and the Ph.D. degree in computer science from the City University of Hong Kong, SAR, China, in 2019. 

He is currently an associate professor with the School of Computer Science and Engineering, Southeast University, China. His research interests include machine learning, Bayesian method, spectral clustering and low-rank modeling.
\end{IEEEbiography}
\vspace{-10 mm}
\begin{IEEEbiography}[
 {
  \includegraphics[width=1in,height=1.75in,clip,keepaspectratio]{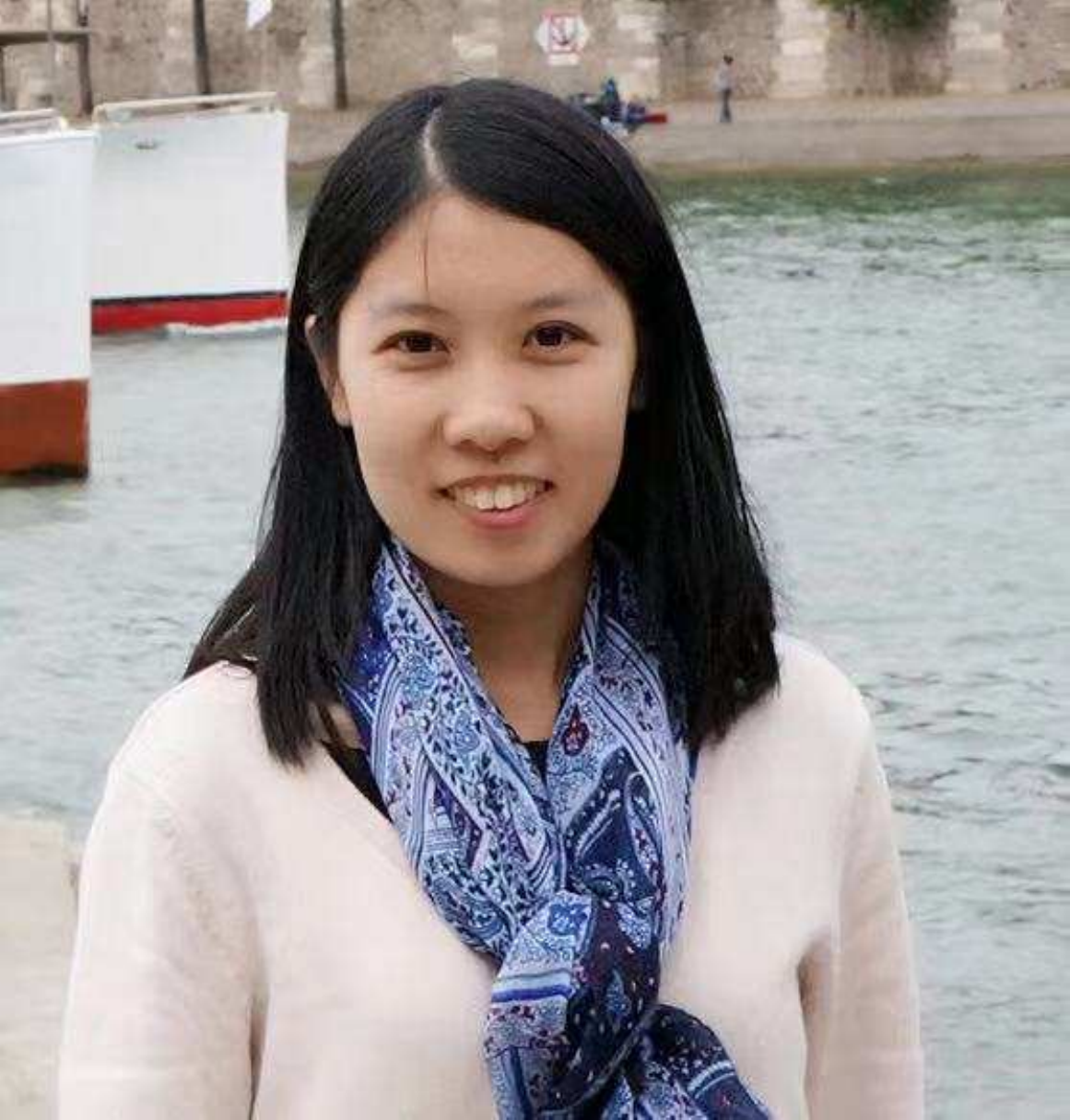}
 }]{Hui Liu}
received the B.Sc. degree in communication engineering from Central South University, Changsha, China, the M.Eng. degree in computer science from Nanyang Technological University, Singapore, and the Ph.D. degree from the Department of Computer Science, City University of Hong Kong, Hong Kong. From 2014 to 2017, she was a Research Associate at the Maritime Institute, Nanyang Technological University. She is currently an Assistant Professor with the School of Computing Information Sciences, Caritas Institute of Higher Education, Hong Kong. Her research interests include image processing and machine learning. 
\end{IEEEbiography}

\vspace{-10 mm}

\begin{IEEEbiography}[{\includegraphics[width=1in,height=1.75in,clip,keepaspectratio]{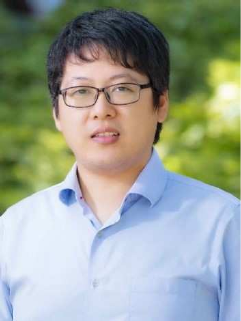}}]{Junhui Hou}
(Senior Member) is an Assistant Professor with the Department of Computer Science, City University of Hong Kong. He received the B.Eng. degree in information engineering (Talented Students Program) from the South China University of Technology, Guangzhou, China, in 2009, the M.Eng. degree in signal and information processing from Northwestern Polytechnical University, Xian, China, in 2012, and the Ph.D. degree in electrical and electronic engineering from the School of Electrical and Electronic Engineering, Nanyang Technological University, Singapore, in 2016. His research interests fall into the general areas of multimedia signal processing, such as image/video/3D geometry data representation, processing and analysis, graph-based clustering/classification, and data compression.

He received the Chinese Government Award for Outstanding Students Study Abroad from China Scholarship Council in 2015 and the Early Career Award (3/381) from the Hong Kong Research Grants Council in 2018. He is an elected member of IEEE MSA-TC, IEEE VSPC-TC, and IEEE MMSP-TC. He is currently an Associate Editor for IEEE Transactions on Image Processing, IEEE Transactions on Circuits and Systems for Video Technology, Signal Processing: Image Communication, and The Visual Computer. He also served as the Guest Editor for the IEEE Journal of Selected Topics in Applied Earth Observations and Remote Sensing and Journal of Visual Communication and Image Representation, and as an Area Chair of ACM MM’19-22, IEEE ICME’20, VCIP’20-22, ICIP’22, MMSP’22, and WACV’21.
\end{IEEEbiography}

\end{document}